\pdfoutput=1

\documentclass[11pt]{article}

\usepackage[preprint]{acl}

\usepackage{times}
\usepackage{latexsym}
\usepackage{paralist}
\usepackage{marvosym}

\usepackage[T1]{fontenc}

\usepackage[utf8]{inputenc}

\usepackage{microtype}

\usepackage{inconsolata}

\usepackage{graphicx}
\usepackage{multicol}
\usepackage{multirow}
\usepackage{utfsym}
\usepackage{makecell}
\usepackage{fontawesome}
\usepackage{amsmath}
\usepackage{booktabs}

\usepackage[ruled,vlined]{algorithm2e}

\newcommand{\ie}{\textit{i.e.}}
\newcommand{\eg}{\textit{e.g.}}
\newcommand{\hytt}[1]{\texttt{\hyphenchar \font=\defaulthyphenchar #1}}

\newcommand{\github}{\raisebox{-1.5pt}{\includegraphics[height=1em]{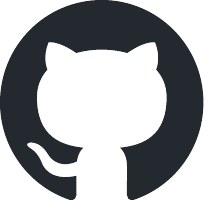}}}
\newcommand{\pjpage}{\raisebox{-1.5pt}{\includegraphics[height=1em]{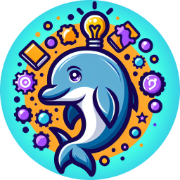}}}
\title{\textsc{Dolphin}: Moving Towards Closed-loop Auto-research through Thinking, Practice, and Feedback}

\author{
Jiakang Yuan$^{\clubsuit,\spadesuit}$, Xiangchao Yan$^{\spadesuit}$, Shiyang Feng$^{\clubsuit,\spadesuit}$, \textbf{Bo Zhang}$^{\spadesuit,\ddagger,}$\textsuperscript{\Letter}, Tao Chen$^{\clubsuit,}$\textsuperscript{\Letter}, \\ \textbf{Botian Shi}$^{\spadesuit}$,  \textbf{Wanli Ouyang}$^{\spadesuit}$, \textbf{Yu Qiao}$^{\spadesuit}$, \textbf{Lei Bai}$^{\spadesuit,}$\textsuperscript{\Letter}, \textbf{Bowen Zhou}$^{\spadesuit}$ \\ [1ex]
$^{\clubsuit}$Fudan University, $^{\spadesuit}$Shanghai Artificial Intelligence Laboratory \\ [1ex]
{\pjpage\ \texttt{\url{https://alpha-innovator.github.io/Dolphin-project-page/}}} \\
{\github\ \texttt{\url{https://github.com/Alpha-Innovator/Dolphin}}} \\
}

\begin{document}
\maketitle

\newcommand\blfootnote[1]{%
\begingroup
\renewcommand\thefootnote{}\footnote{#1}%
\endgroup
}

\blfootnote{{\Letter \ Corresponding Authors}, $\ddagger$~Project Lead \\}

\begin{abstract}
    The scientific research paradigm is undergoing a profound transformation owing to the development of Artificial Intelligence (AI). Recent works demonstrate that various AI-assisted research methods can largely improve research efficiency by improving data analysis, accelerating computation, and fostering novel idea generation. To further move towards the ultimate goal (\textit{i.e.}, automatic scientific research), in this paper, we introduce \textsc{Dolphin}, a closed-loop LLM-driven framework to enhance the automation level of scientific research. 
    \textsc{Dolphin} first generates novel ideas based on feedback from previous experiments and relevant papers ranked by the topic and task attributes. Then, the generated ideas can be implemented using a code template refined and debugged with the designed exception-traceback-guided local code structure. Finally, \textsc{Dolphin} automatically analyzes the results of each idea and feeds the results back to the next round of idea generation. Experiments are conducted on the benchmark datasets of different topics and a subset of MLE-bench. Results show that \textsc{Dolphin} can continuously improve the performance of the input topic in a loop. We highlight that \textsc{Dolphin} can automatically propose methods that are \textbf{comparable to the state-of-the-art} in some tasks such as 3D point classification.
\end{abstract}
\section{Introduction}
\label{sec:intro}

The rapid evolution of AI~\cite{achiam2023gpt4,Anthropic2024claude3,yang2024qwen2,dubey2024llama3} has profoundly transformed various fields
~\cite{yao2024lawyer,yan2025surveyforge,luo2022biogpt}, accelerating scientific research processes like scientific data collection and processing \cite{han2024cra5,yan2025surveyforge}, computation \cite{jumper2021highly,chen2023fengwu}, and innovation \cite{qi2024large}. Under this trend, the research paradigm is shifting from completely human-driven research to AI-assisted research~\cite{assafelovic2023gptresearcher}. More recently, the continuous upgrading of LLMs has promoted the evolution of AI-assisted research to automatic scientific research~\cite{AIScientist,si2024stanfordcan}.

\begin{figure}[t]
\vspace{-8pt}
  \centering
   \includegraphics[width=1.0\linewidth]{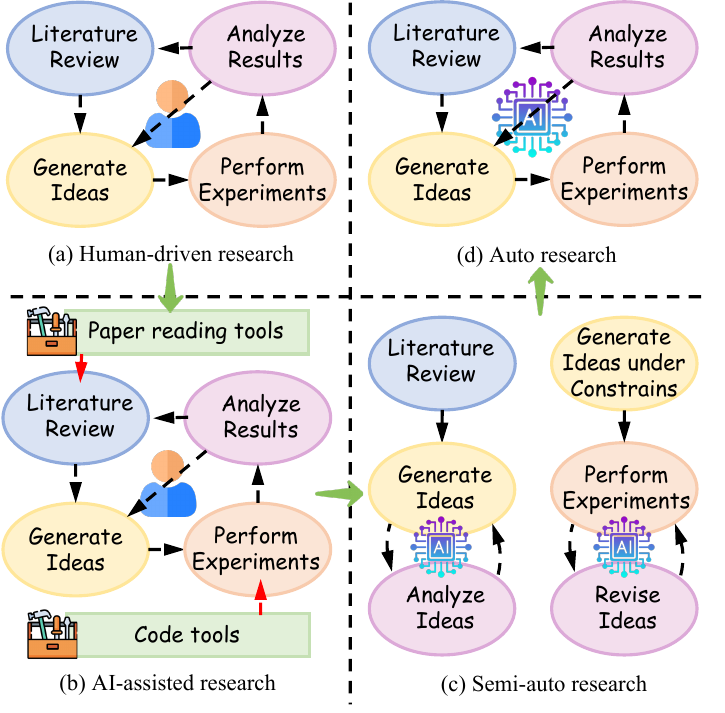}
   \vspace{-15pt}
   \caption{Comparisons of the four stages in the evolutionary trajectory towards auto-research including (a) Entirely human-driven research, (b) AI-assisted research, (c) Semi-automatic research, and (d) Auto-research.}
   \label{fig:fig1}
\vspace{-5pt}
\end{figure}


The evolutionary trajectory from human-driven research to automatic research consists of four stages as shown in Fig.~\ref{fig:fig1}. \textbf{1)} \textit{The entirely human-driven stage} requires humans to complete all aspects including idea generation and experiments. \textbf{2)} \textit{In the AI-assisted research stage}, researchers use LLMs-based tools~\cite{assafelovic2023gptresearcher,copilot} to improve the research efficiency. For example, GPT-researcher~\cite{assafelovic2023gptresearcher} can help us to decompose complex tasks and generate research reports using LLMs. \textbf{3)} \textit{The semi-automatic research stage} enables automation in certain processes of scientific research. For example, recent works~\cite{si2024stanfordcan,li2024chain,su2024two} leverage LLMs to automatically generate ideas for different topics by drawing on relevant works. \textbf{4)} The fourth stage of AI-promoting scientific research is \textit{auto-research stage}, where AI automatically handles the entire research process from conception to completion. Recently, AI-Scientist~\cite{AIScientist} introduced an open automatic research framework automating key tasks including idea generation, experimental verification, and academic paper writing.

Despite the encouraging progress made in existing works, auto-research still faces key challenges:

\textbf{First}, accurately assessing the effectiveness of AI-generated ideas remains a major hurdle. Most studies~\cite{si2024stanfordcan,li2024chain,su2024two} rely on either human evaluation or LLMs to evaluate the quality of generated ideas. However, \textit{merely focusing on the novelty of an idea itself does not adequately reflect its effectiveness in experimental validation}. While works like AI-Scientist~\cite{AIScientist} incorporate experimental validation, they use simple, the self-constructed datasets, limiting meaningful comparisons with existing methods in the same field. 

\textbf{Second}, another limitation of prior works~\cite{si2024stanfordcan,li2024chain,su2024two} lies in the absence of a feedback mechanism between the experimental validation and the idea generation \textendash\ a process that is fundamental to human research. Human researchers refine their ideas and approaches iteratively based on experimental outcomes, which serves as a crucial pathway for improving the quality of research ideas.

To address these challenges and facilitate further progress towards automatic scientific research, in this work, we propose \textsc{Dolphin}, a closed-loop auto-research framework which is composed of three key stages in the research cycle (\ie, idea generation, experimental verification, and results feedback). In each research loop, \textsc{Dolphin} first generates ideas based on the previous experimental results and the retrieved relevant papers which are filtered by judging the topic relevance and task attribute relevance between the retrieved papers and the input topic. Then, the proposed idea can be further implemented using a code template refined and debugged by the designed traceback-guided debugging module, which analyzes the local code structure related to the error-traceback to streamline the debugging process. Finally, \textsc{Dolphin} automatically analyzes the outcomes of successfully executed experiments, providing guidance for the next iteration of idea generation.



To experimentally validate the generated scientific ideas, we selected commonly-used public benchmarks across different tasks and modalities such as ModelNet40~\cite{wu2015modelnet}, CIFAR-100~\cite{krizhevsky2009cifar}, and SST-2~\cite{socher2013recursive}. The results demonstrate that \textsc{Dolphin} generates ideas that outperform the selected baselines like PointNet~\cite{qi2017pointnet}, WideResNet~\cite{he2016resnet}, and BERT-base~\cite{devlin2019bert}. We were \textbf{supervised} to observe that \textsc{Dolphin} is capable of proposing more concise method implementations while surpassing the performance of current human-designed state-of-the-art approaches in certain fields, as reported in Tab.~\ref{tab:case_study}. Besides, experiments on tasks from MLE-bench reveal that \textsc{Dolphin} can integrate with code generation pipelines like AIDE and further support version updates for technologies or code.


Our contribution can be summarized as follows:
\begin{compactitem}
    \item We propose \textsc{Dolphin}, a closed-loop auto-research framework covering three key stages in the research cycle, including generating ideas, performing experiments, and feedback.
    \item To improve the efficiency of auto-research, we propose task-attribute-guided paper ranking and exception-traceback-guided debugging process to improve the quality of generated ideas and the successful rate of code execution, respectively.
    \item Experimental results on benchmark datasets show that \textsc{Dolphin} can generate high-quality ideas and conduct validations in the loop. 
\end{compactitem}
\section{Related Works}
\label{sec:related}

\begin{figure*}[t!]
\vspace{-12pt}
  \centering
   \includegraphics[width=0.96\linewidth]{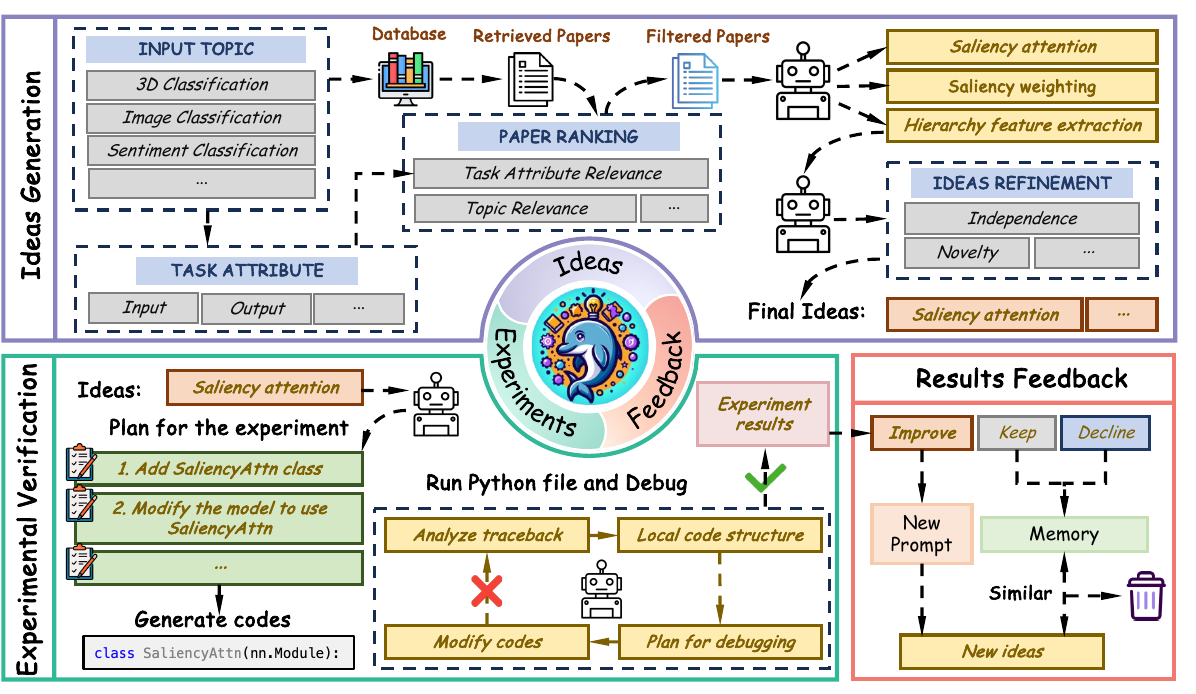}
   \vspace{-8pt}
   \caption{\textsc{Dolphin} first generates a set of ideas based on the retrieved papers. After filtering ideas, experimental plans will be generated for these filtered ideas. Then, codes can be generated and debugged using the proposed error-traceback-guided debugging process. Finally, the results of successfully executed experiments will be auto-analyzed and reflected into the next round of ideas generation.}
   \label{fig:framework}
\vspace{-7pt}
\end{figure*}


\vspace{-2pt}
\subsection{Open-ended Scientific Research}
\vspace{-2pt}

Recent studies~\citep{qi2023large, Wang2023SciMONSI, Yang2023LargeLM, qi2024large, AIScientist, hu2024nova} have demonstrated that Large Language Models (LLMs) have the potential to generate novel research ideas, a finding that has sparked widespread discussion in academia. 
\citet{li2024chain} generate ideas based on the analysis of the paper chain. \citet{si2024stanfordcan} find that LLM-generated ideas are innovative but their feasibility needs improvement. Some works~\citep{Wang2023SciMONSI, hu2024nova} improve the novelty of ideas using iterative optimization strategies. In addition, several works~\citep{qiu2023phenomenal,zhou2024hypothesis} try to use LLMs to generate hypotheses. For example, \citet{yang2023large} generate hypotheses from web corpus and ~\citet{wang2023scimon} generate hypotheses from literature. However, these works lack empirical verification of their practical effectiveness. Recently, AI-Scientist~\citep{AIScientist} and 
Agent Laboratory~\citep{schmidgall2025agent} introduce the end-to-end framework that automates the process from idea generation to experimental execution and paper writing. However, the experimental validation of its idea generation remains limited, lacking evaluation conducted on real-world datasets or scenarios. Furthermore, the framework lacks a feedback mechanism that connects experimental validation to idea generation, unlike human researchers who iteratively refine their hypotheses based on experimental outcomes.

\vspace{-2pt}
\subsection{Scope-limited Scientific Research.}
\vspace{-2pt}

Several studies have effectively applied LLMs to specific scientific discovery tasks. For example, AutoML-GPT~\citep{zhang2023automl} leverages LLMs for hyperparameter tuning by combining model and data descriptions as prompts to predict training logs. AgentHPO~\citep{liu2024large} introduces a creator-executor framework to iteratively optimize hyperparameters based on historical trials. Similarly, MLCopilot~\citep{zhang2023mlcopilot} constructs an experience pool from historical data to enable LLMs-based hyperparameter prediction. EvoPrompting~\citep{chen2024evoprompting} improves the in-context prompting examples of LLMs to achieve effective code-level neural architecture design. In contrast to these scope-limited approaches, our method focuses on more open-ended scientific discovery, spanning from idea generation to experimental validation, and achieving a fully closed-loop research lifecycle.


\vspace{-2pt}
\section{Methods}
\label{sec:method}
\vspace{-2pt}

In this section, we introduce \textsc{Dolphin}, a closed-loop auto-research framework as shown in Fig.~\ref{fig:framework}, which is mainly composed of \textbf{an idea generation process}, \textbf{an experiments verification process}, and \textbf{a result feedback process}. The closed-loop means that the experimental results will be fed back into the idea generation process and the above three processes form a research cycle. In the idea generation process, \textsc{Dolphin} generates ideas based on the feedback and retrieved papers that are filtered by a designed task-attribute-guided paper ranking process. Then, \textsc{Dolphin} formulates experimental plans and proceeds to generate and debug code using a specifically designed error-traceback-guided debugging process. Finally, the results will be analyzed and utilized as feedback for the next cycle of ideas generation. The following sections detail the ideas generation process, experiments verification process, and results feedback process in Sec.~\ref{sec:method_idea}, Sec.~\ref{sec:method_exp}, and Sec.~\ref{sec:method_result} respectively.

\vspace{-2pt}
\subsection{Ideas Generation Process}
\label{sec:method_idea}
\vspace{-2pt}

\textit{``A good beginning is half done.''} As the beginning of the research cycle, high-quality ideas are crucial to the entire research. A promising approach to generating high-quality ideas is to imitate the behaviors of human researchers. They typically first conduct literature reviews and then generate ideas based on the literature~\cite{randolph2019guide} and previous experimental experience (detail in Sec.~\ref{sec:method_result}). \textsc{Dolphin} typically divided the idea generation process into two steps including \textbf{1)} paper retrieval and ranking, and \textbf{2)} ideas generation and filtering. 

\noindent \textbf{Paper Retrieval and Ranking.} To generate high-quality ideas, the first step is to retrieve papers that are relevant to the topic. Given a research topic (\eg, 3D classification), \textsc{Dolphin} begins by searching for relevant papers using Semantic Scholar API\footnote[1]{\url{https://www.semanticscholar.org/product/api}\label{fn:1}}, obtaining the essential information such as titles and abstracts. However, the initially retrieved papers are not always directly related to the input topic, which can limit their usefulness in generating subsequent ideas. For instance, if the input topic is 3D classification, some retrieved papers might pertain to 3D detection~\cite{wang2024autosurvey}. Although these topics are interconnected, they typically focus on different challenges. As a result, it is necessary to filter out papers that are irrelevant to the specific topic.

To this end, we design a task-attribute-guided paper ranking process that aims to assign a higher score to the paper relevant to the input topic and task attribute. In detail, \textsc{Dolphin} ranks the retrieved papers based on two main criteria: \textbf{1)} relevance to the input topic, and \textbf{2)} alignment of task attributes with those of the input topic. The task attributes typically define a task, including model inputs, model outputs, and other characteristics. Specifically, the LLM is first utilized to extract the task attributes of the input topic and then prompt it to score (\ie, 1-10) each retrieved paper based on the designed criteria. The detailed prompts can be found in our appendix. After scoring, \textsc{Dolphin} filters out papers with scores below 8, using the remaining papers as the references of the subsequent ideas generation process.

\noindent \textbf{Ideas Generation and Filtering.} After obtaining retrieved papers, the next step is to generate ideas based on the references. \textsc{Dolphin} begins by prompting the LLM to generate $N$ novel and non-redundant ideas, each comprising a title, a brief experiment plan, and a summary. However, these generated ideas are not consistently novel, and some ideas are similar to one another. As a result, directly performing experiments on such ideas will cost substantial time and computational resources, further reducing research efficiency. To address this, we introduce a further idea-filtering procedure that filters out non-novel or redundant ideas.

To be specific, \textsc{Dolphin} first examines the independence of ideas to ensure non-redundancy. Given $N$ ideas $[I_1, I_2, ..., I_N]$, \textsc{Dolphin} first extract the sentence-level embedding $[f_1, f_2, ..., f_N]$ based on the summary of each ideas. Then, an idea bank $\mathbf{B}$ is constructed to store embeddings of the ideas that have been checked to be independent. $\mathbf{B}$ is initialized as empty in the first loop and initialized with previous ineffective ideas in the following loops which will be introduced in Sec.~\ref{sec:method_result}. When examining the $i$-th idea, \textsc{Dolphin} calculates its cosine similarity with existing ideas stored in $\mathbf{B}$ and discards the current idea when the max similarity exceeds a predetermined threshold (\ie, 0.8). After filtering out redundant ideas, the remaining $N'$ ideas will undergo the novelty check. Following AI-Scientist~\cite{AIScientist}, \textsc{Dolphin} simply prompt the LLMs to decide whether the idea is novel based on the searched papers by Semantic Scholar API\footref{fn:1}. Only ideas identified as novel and independent will proceed to the subsequent experimental verification process.

\vspace{-2pt}
\subsection{Experimental Verification Process}
\label{sec:method_exp}
\vspace{-2pt}

\textit{``Jump in and get your feet wet.''} Experimental verification is crucial in the research cycle as it serves as the most effective way to confirm the effectiveness of proposed ideas. Most existing works~\cite{li2024chain,su2024two} assess the novelty of AI-generated ideas via LLMs or humans. However, without closed-loop experimental verification, the effectiveness of these ideas remains uncertain. In contrast, \textsc{Dolphin} screens out truly effective ideas through an experimental verification process.

Given an idea generated by the ideas generation process (Sec.~\ref{sec:method_idea}) and reference codes, \textsc{Dolphin} first prompts the LLM to generate detailed experimental plans and then modifies the reference codes according to the idea and the generated plans. After modified codes, the experiment will automatically proceed. However, \textit{we find that the execution success rate is relatively low} since LLMs encounter significant challenges in modifying code with complicated nested relationships (\eg, between \texttt{class} and \texttt{function}), while ensuring complete error-free execution. This will further reduce the efficiency of verifying ideas and research.

\begin{figure}[t]
  \centering
   \includegraphics[width=0.90\linewidth]{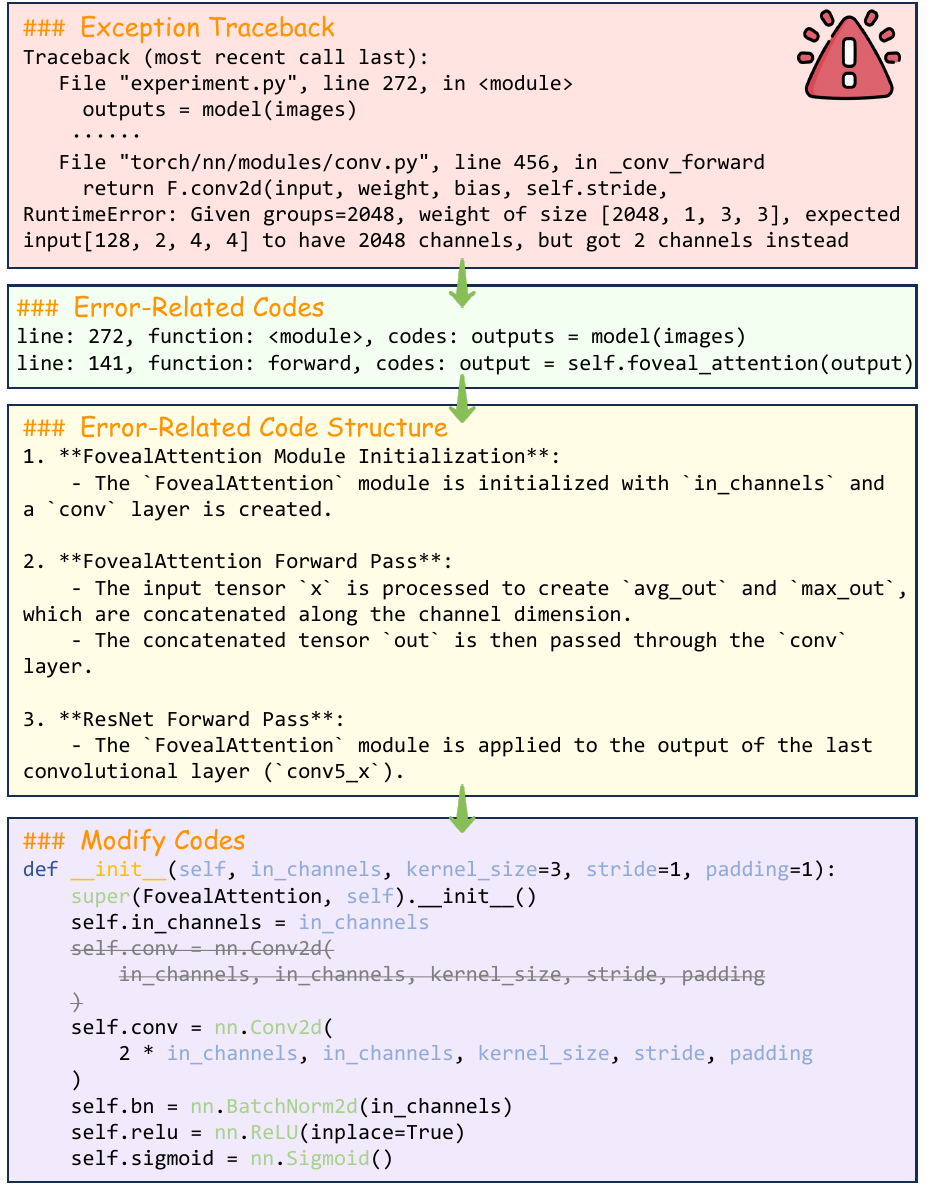}
    \vspace{-5pt}
   \caption{Debugging with traceback-guided local code structure.}
   \label{fig:debug}
\vspace{-12pt}
\end{figure}

Based on our observation, we design an exception-traceback-guided debugging process as shown in Fig.~\ref{fig:debug}, aiming at assisting the LLMs in comprehending code logic with local code structure. Specifically, to generate the code structure related to the code errors, \textsc{Dolphin} first extracts information in exception traceback, including function name, line, and code, since traceback contains the nested information between functions. Note that \textsc{Dolphin} only focuses on custom codes, excluding library function calls. Then, \textsc{Dolphin} prompts the LLM to generate the code structure under the guidance of extracted exception traceback information. After that, the LLM analyzes the exception traceback and local code structure to make necessary modifications, enabling automatic code execution after these adjustments. The debugging process will be repeated until either successful execution is achieved or the predetermined maximum number of debugging times is reached. Finally, all successfully implemented ideas will undergo comprehensive analyses in the next phase.

\vspace{-2pt}
\subsection{Results Feedback Process}
\label{sec:method_result}
\vspace{-2pt}

\textit{``Experience is the best teacher.''} Human researchers often analyze experimental results to further propose new ideas or improve existing ideas, since insights from previous experiments can be leveraged to effectively enhance the quality of subsequent idea generation. However, recent works either implement feedback mechanisms within the isolated idea generation process~\cite{li2024chain} or lack feedback mechanisms entirely~\cite{AIScientist}. To address this limitation, \textsc{Dolphin} analyzes experimental results and incorporates the findings into the subsequent round of ideas generation process.

\textsc{Dolphin} first divides the experimental results into three categories (\ie, \textbf{improvement}, \textbf{maintenance}, and \textbf{decline}) compared to the performance of the reference codes. Our goal is to discourage the development of ideas that have previously led to stagnant or declining performance, while actively promoting the creation of innovative concepts or iterations based on past ideas that enhance the model performance. In detail, \textsc{Dolphin} incorporate the embeddings of summaries from ideas that maintain or improve the performance into $\mathbf{B}$ defined in Sec.~\ref{sec:method_idea}. In this way, ideas will be filtered out if they are similar to previous ideas that cannot improve the performance and avoid redundant verification of the ineffective ideas. Besides, \textsc{Dolphin} incorporates performance-enhancing ideas into the idea generation prompt for the next loop. Detailed prompts are shown in Appendix~\ref{sec:dolphin_prompt}.


\begin{table*}[ht]
\vspace{-8pt}
\centering
\setlength\tabcolsep{15pt}
\resizebox{0.98\linewidth}{!}{
\begin{tabular}{c|cc|c|c}
\toprule
\multirow{2}{*}{\textbf{Tasks} }  & \multicolumn{2}{c|}{ \textbf{ModelNet40} }  &  \textbf{CIFAR-100}  &  \textbf{SST-2}  \\
& \textbf{OA (\%)} &  \textbf{mAcc. (\%)}  &  \textbf{Top-1 Acc. (\%)}  &   \textbf{Acc. (\%)}  \\
\hline
Baseline  &  89.2 (PointNet) & 86.2 (PointNet)  &  79.5 (WRN-28-10) &  -  \\
Baseline$^\dagger$  &  91.0 (PointNet) &  87.6 (PointNet)  &  81.2 (WRN-28-10) &  91.0 (BERT-base) \\
Avg. Improvement  &  92.0 (+1.0) & 88.7 (+1.1)  &  81.8 (+0.6)  &  91.8 (+0.8) \\
Max Improvement  &  93.9 (+2.9) & 91.1 (+3.5)  &   82.0 (+0.8)  &  92.5 (+1.5) \\
Human designed  &   93.8 (GPSFormer) & 91.8 (GPSFormer)  &   82.2 (ResNeXt)   &   93.1 (BERT-large)  \\
Number ideas    &   5 / 40   &    5 / 40   &    6 / 40   &   6 / 40  \\
\bottomrule
\end{tabular}
}
\vspace{-6pt}
\caption{Experimental verifications on 3D point classification, 2D image classification, and sentiment classification tasks. Number ideas refers to the number of ideas that can achieve performance gains. $\dagger$ denotes the results of our implementation. Avg. Improvement and Max Improvement represent the average and maximum improvement of all ideas that can improve the baseline performance.}
\label{tab:main}
\vspace{-12pt}
\end{table*}

\section{Experiment}
\label{sec:exp}

\vspace{-2pt}
\subsection{Experimental Setups}
\vspace{-4pt}

\noindent \textbf{Tasks and Datasets.} We conduct auto-research on three topics including image classification, 3D classification, and sentiment classification. \textit{For 2D classification task}, we evaluate our method on CIFAR-100~\cite{krizhevsky2009cifar}, which is widely-used in computer vision. \textit{For 3D classification task}, we use ModelNet40~\cite{wu2015modelnet} which is a 3D CAD dataset and consist of 40 categories. \textit{For the sentiment classification task}, we use Stanford Sentiment Treebank (SST-2) dataset~\cite{socher2013recursive}. We also conduct experiments on MLE-bench~\cite{chan2024mle}, which is designed to evaluate the capability of models in handling real-world ML tasks on Kaggle competitions. Further details of datasets can be found in Appendix~\ref{task}.

\noindent \textbf{Implementation Details.} For the ideas generation process, we use \hytt{gpt-4o-2024-08-06}~\cite{openai2024gpt4o} as our LLM agent. The total number of retrieved papers is set to 50 and only papers with scores higher than 8 will be treated as references for the ideas generation process. We generate 20 ideas in each loop and the threshold of the independence filtering is set to 0.8. We use \hytt{sentence-transformer/all-roberta-large-v1}~\cite{reimers-2020-multilingual-sentence-bert} to extract the summary embedding of each idea. For the experimental verification process, we use \hytt{deepseek-v2.5}~\cite{zhu2024deepseek} deployed by ollama~\cite{ollama} as our code agent. The maximum number of debugging attempts is set to 5 for experimental efficiency. Following AI-Scientist~\cite{AIScientist}, we use self-reflection to first eliminate some syntax errors before executing the program and use aider as the framework to call LLM agents. The same hyper-parameters and models employed in the ideas generation process are utilized in the results feedback.

\vspace{-2pt}
\subsection{Main Results}
\vspace{-2pt}

We evaluate \textsc{Dolphin}'s capabilities across various tasks covering point clouds, images, and language modalities. In this section, we conduct experiments on each task for two loops (\ie, 40 ideas). 

\noindent \textbf{Results on 3D Point Classification.}~We conduct experiments on 3D classification task using PointNet~\cite{qi2017pointnet} as our baseline model. 
As illustrated in Tab.~\ref{tab:main}, a total of 5 ideas achieve performance gains in two loops, with an average improvement of 1.0\% OA, which shows that \textsc{Dolphin} can generate and verify effective ideas. Besides, the maximum improvement can achieve 93.9\% OA, which largely improves the performance compared with the human-designed baseline (\ie, 91.0\% achieved by PointNet). More excitingly, such results achieved by auto-research obtain comparable performance to the current SoTA method (\ie, 93.8\% achieved by GPSFormer~\cite{wang2024gpsformer}). This method is carefully designed by human researchers based on Transformer architecture. We would like to emphasize that for a fair comparison, we compare the 3D methods without pre-training and the voting mechanism.

\noindent \textbf{Results on 2D Image Classification.}~Further, we conduct experiments on the image classification task, using WRN-28-10~\citep{zagoruyko2016wrn} as our baseline model. As shown in Tab.~\ref{tab:main}, the average improvement and max improvement are 0.6\% and 0.8\%, respectively. Notably, the idea generated and performed automatically by \textsc{Dolphin} can obtain comparable performance to hand-crafted methods such as ResNeXt~\citep{liu2022convnet} (\eg, 82.0\% compared to 82.2\%). It should be noted that Transformer-based methods such as ViT~\citep{dosovitskiy2020image} are not included in our comparison, due to their heavy dependence on large-scale pre-training, which, at this stage, requires significant resources to validate their effectiveness.

\noindent \textbf{Results on Sentiment Classification.} To verify the effectiveness of \textsc{Dolphin} on different modalities, we also perform experiments on sentiment classification task. We conduct experiments on SST-2~\cite{socher2013recursive} and report the classification accuracy. In Tab.~\ref{tab:main}, we fine-tune the pre-trained BERT-base~\cite{devlin2019bert} as our baseline model. It can be seen that \textsc{Dolphin} also generates and performs effective ideas (\eg, 1.5\% improvement on SST-2) on sentiment classification, reducing the performance gap between BERT-base (\ie, 92.5\%) and BERT-large (\ie, 93.1\%).

\noindent \textbf{Results of tasks in MLE-bench.} We further conduct experiments on tasks from MLE-bench~\cite{chan2024mle} as shown in Tab.~\ref{tab:mlebench}. First, \textsc{Dolphin} can flexibly integrate with code generation frameworks such as AIDE~\cite{schmidt2024aide} and Agent Laboratory~\cite{schmidgall2025agent}. For example, based on the code template generated by AIDE, \textsc{Dolphin} further improves the performance on tasks such as insult detection (\ie, 81.0\% $\rightarrow$ 84.7\%) and achieves \textbf{gold medal} level performance on Kaggle. Additionally, \textsc{Dolphin} performs version updates for technologies or code. For example, based on previous code from Kaggle, \textsc{Dolphin} can achieve further accuracy breakthroughs (\eg, 95.3 $\rightarrow$ 96.2 on tabular playground).

\begin{table}[t!]
\vspace{-8pt}
\centering
\setlength\tabcolsep{6pt}
\resizebox{1.0\linewidth}{!}{
\begin{tabular}{ccccc}
\toprule
\textbf{Tasks}   & \textbf{Domain}  &  \textbf{Code src.}  &  \textbf{Prev. score}  &  \textbf{Score} \\
\hline
\makecell[c]{Detecting insults in\\[-3.5pt] social commentary} & \faFileTextO & AIDE & 81.0 & 84.7 \\
\makecell[c]{Tabular playground \\[-3pt] series dec 2021} & \faTable & Kaggle & 95.3 & 96.2 \\
\makecell[c]{Jigsaw toxic \\[-3.5pt] comment classification	} & \faFileTextO & Kaggle & 94.7 & 97.2 \\ \bottomrule

\end{tabular}
}
\vspace{-7pt}
\caption{Results of tasks in MLE-bench. Code src. denotes the source of the code template (AIDE: auto-generate by AIDE, Kaggle: Kaggle public code). }
\label{tab:mlebench}
\vspace{-2pt}
\end{table}



\begin{table}[t!]
\centering
\setlength\tabcolsep{9pt}
\resizebox{0.98\linewidth}{!}{
\begin{tabular}{ccc}
\toprule
\textbf{Method} &  \textbf{Novelty}   &   \textbf{Cost} (Avg.) \\
\hline
Naive generation & 8 / 20 & \$0.106 \\
Generation with naive retrieval & 13 / 20 &  \$0.187  \\
Ours (task attribute filtering) & 19 / 20 &  \$0.184  \\
\bottomrule

\end{tabular}
}
\vspace{-6pt}
\caption{Results of ideas generation process. The novelty is evaluated by \hytt{gpt-4o-2024-08-06}. Cost (Avg.) is the cost per idea including paper retrieval, ideas generation, and novelty check.}
\label{tab:ideas}
\vspace{-10pt}
\end{table}

\vspace{-4pt}
\subsection{Further Analyses}
\label{ablation}
\vspace{-4pt}

\noindent \textbf{Analysis on Ideas Generation Process.} To evaluate the effectiveness of the ideas generation process, we conduct further ablation studies by comparing naive generation, generation with retrieved papers, and our proposed method. Naive generation refers to the direct use of LLMs to generate ideas based on the seed idea and reference codes, similar to the approach used by AI-Scientist~\cite{AIScientist}. Generation with retrieved papers involves directly searching papers based on the topic and filtering them by their relevance to the input topic. As shown in Tab.~\ref{tab:ideas}, the naive generation yields the poorest results, with more than half of the ideas being judged as not novel. Furthermore, the quality of generated ideas is significantly improved when using naive retrieved papers, as this approach more closely aligns with the way human researchers generate ideas. However, as mentioned in Sec.~\ref{sec:method_idea}, this approach tends to retrieve irrelevant papers and will mislead LLMs. As indicated in the first line of Tab.~\ref{tab:frequency}, some papers primarily focus on 3D detection or point cloud completion, where the design approach of the model is entirely different from that of point classification. This phenomenon can be well handled by \textit{the designed paper ranking process}. As illustrated in Tab.~\ref{tab:ideas} and Tab.~\ref{tab:frequency}, the number of novel ideas significantly increased from 8/20 to 19/20, while the proportion of papers related to irrelevant topics substantially decreased. This improvement can be attributed to the availability of more relevant reference papers during the ideas generation process. Note that the higher occurrence of the keyword ``segmentation'' is due to that, many studies concurrently perform both point classification and segmentation tasks. In addition, the average cost per idea is shown in Tab.~\ref{tab:ideas}. It can be seen that the cost of each idea is very small. Besides, the cost of generating each idea is relatively higher when retrieving papers. This is mainly due to the retrieval process and the longer prompt required for idea generation (\ie, both the title and abstract are fed into the LLMs as references).

\begin{table}[t!]
\vspace{-8pt}
\centering
\setlength\tabcolsep{5pt}
\resizebox{1.0\linewidth}{!}{
\begin{tabular}{ccccc}
\toprule
\textbf{Keywords}  &  \textbf{Classification}  & \textbf{Detection}  & \textbf{Segmentation}  &  \textbf{Completion}  \\
\hline
Naive & 82 & 17 & 38 & 16  \\
Filter (ours) & 109 & 4 & 43 & 0 \\
\bottomrule

\end{tabular}
}
\vspace{-7pt}
\caption{For the 3D classification task, the frequency of each keyword is determined from the retrieved papers, focusing only on those words that appear in the abstracts and titles of papers scoring above 8 points in the ranking process. ``Naive'' and ``Filter'' refer to naive retrieval and filtering based on task attributes.}
\label{tab:frequency}
\vspace{-10pt}
\end{table}

\noindent \textbf{Analysis on Experimental Verification Process.} The success rate of experiment execution is crucial for improving research efficiency. We further conducted studies on the experimental verification process. As illustrated in Tab.~\ref{tab:execution}, we conduct experiments on three approaches: \textbf{1)} directly feed the exception traceback to LLM, \textbf{2)} extract the local code structure based on exception traceback, and then feed the local code structure and traceback to LLM, and \textbf{3)} extract local code structure according to the information derived from the exception traceback, and then feed the local code structure and traceback to LLM. 

Firstly, we find that the successful execution rate is relatively low (\eg, 4 / 15) when directly feeding the traceback into LLM for debugging since the LLM cannot fully understand the complicated nested relationships in the codes. For example, when a dimension mismatch error occurs between networks and feature dimensions, LLMs can easily locate where the error occurs. However, since the feature might be obtained through multiple nested modules, LLMs fail to correct the network dimension. Then, by adding the local code structure according to exception traceback, the success rate does not significantly improve. This is because the exception traceback contains lots of information about called libraries, which makes LLMs generate code structures irrelevant to our custom codes. Further, by guiding LLMs to generate code structures with information extracted from traceback, the execution rate can be largely improved (\ie, 33.3\%$\rightarrow$50.0\%). This is due to the extracted information containing custom code information related to the exception, enabling LLMs to focus on relevant functions and variables. Note that to improve efficiency, we only allow a maximum of 5 debugging iterations in our experiments.

\begin{table}[t]
\vspace{-8pt}
\centering
\setlength\tabcolsep{6pt}
\resizebox{0.80\linewidth}{!}{
\begin{tabular}{ccccc}
\toprule
\multirow{2}{*}{ \textbf{L.C.S.} } & \multirow{2}{*}{ \textbf{Traceback} } &  \multicolumn{3}{c}{\textbf{Successful execution}}   \\
& & \textbf{Loop 1} & \textbf{Loop 2} & \textbf{Loop 3}\\
\hline
\usym{2717} & \usym{2717} & 4 / 15 & 5 / 13 & 5 / 14 \\
\usym{2713} & \usym{2717} &  3 / 15 & 5 / 13 & 6 / 14  \\
\usym{2713} & \usym{2713} &  7 / 15 & 6 / 13 & 8 / 14  \\
\bottomrule

\end{tabular}
}
\vspace{-6pt}
\caption{Results of successful execution rate. L.C.S. represents local code structure. Traceback denotes using information extracted from exception traceback. The denominator is the number of ideas after the novelty and independence check.}
\label{tab:execution}
\vspace{-2pt}
\end{table}

\begin{table}[t]
\centering
\setlength\tabcolsep{4pt}
\resizebox{0.90\linewidth}{!}{
\begin{tabular}{ccccc}
\toprule
\textbf{Loop} &  \textbf{Loop 1} & \textbf{Loop 2} & \textbf{Loop 3}  &  \textbf{Total}  \\
\hline
Improvement rate & 2 / 7 & 3 / 6 & 4 / 8 & 9 / 21 \\
Cost (Avg.) & 0.184  & 0.203 & 0.218 & 0.201   \\
\bottomrule

\end{tabular}
}
\vspace{-6pt}
\caption{Performance in different loops. The denominator is the number of successfully executed ideas.}
\label{tab:loop}
\vspace{-5pt}
\end{table}

\begin{table}[t]
\vspace{-8pt}
\centering
\setlength\tabcolsep{4pt}
\resizebox{1.0\linewidth}{!}{
\begin{tabular}{ccc}
\toprule
\textbf{Method}   &   \textbf{Accuracy} (Avg. class)   &   \textbf{Overall accuracy} \\
\hline
\multicolumn{3}{c}{\textit{Human designed methods}} \\
\hline
PointNet~\cite{qi2017pointnet} &86.2 & 89.2\\
PointNet++~\cite{qi2017pointnet++} & - & 91.9 \\
DGCNN~\cite{wang2019dynamic} & 90.2 & 92.9 \\
PointNeXt~\cite{qian2022pointnext} & 90.8 & 93.2 \\
OctFormer~\cite{wang2023octformer} & - & 92.7  \\
GPSFormer~\cite{wang2024gpsformer} & \textbf{91.8 }& 93.8 \\
\hline
\multicolumn{3}{c}{\textit{\textbf{Methods obtained by \textsc{Dolphin}} (auto-research)}} \\ \hline
PointNet-CSR & 91.1 & \textbf{93.9} \\
\bottomrule
\end{tabular}
}
\vspace{-6pt}
\caption{Accuracy on ModelNet40. The results are obtained from 1024 points without voting.}
\label{tab:case_study_sota}
\vspace{-2pt}
\end{table}

\noindent \textbf{Analysis on Results Feedback Process.} To demonstrate the advantages of the closed-loop framework, we further analyze results on different loops. As shown in Tab.~\ref{tab:loop}, we find that the quality of generated ideas without feedback is relatively low. The reasons can be divided into two folds: \textbf{1)} Repeated ideas may be generated without a feedback process, resulting in redundant verification of the same idea and decreased experimental efficiency. \textbf{2)} In the absence of feedback, the model cannot learn what kind of ideas are effective for the specific task. 

In contrast, \textsc{Dolphin} 
effectively solves these challenges through a closed-loop approach, demonstrating progressive improvement in idea quality as the number of iterations increases (\eg, $2/7$ improvement rate in Loop 1 $\rightarrow$ $4/8$ improvement rate in Loop 3). Further, these ideas that improve the performance are different between each round, which further improves the research efficiency and shows the effectiveness of \textsc{Dolphin}. Besides, the average cost slightly 
increases as the iterations continue, since the results will be fed back into the next round of ideas generation process.

\subsection{Case Studies}

\begin{table}[t]
\centering
\setlength\tabcolsep{2pt}
\resizebox{1.0\linewidth}{!}{
\begin{tabular}{cll}
\toprule
\textbf{Diff.} &  \textbf{DGCNN} (Human) &  \textbf{PointNet-CSR} (\textsc{Dolphin}) \\
\hline
\multirow{3}{*}{Idea} & 1) Architecture-level & 1) Module-level \\
 & 2) With learnable parameters & 2) Without learnable parameters \\
 & 3) Repeated blocks & 3) Single module\\
\hline
Impl. & \makecell[l]{Multi-layer Edge with high \\complexity} & \makecell[l]{Single contextual semantic reasoning \\module with low complexity}\\
\hline
\multirow{2}{*}{Results}  & 1) 90.2\% mAcc., 92.9\% OA  & 1) 91.1\% mAcc., 93.9\% OA \\
& 2) {\small $\sim$} 20.86s per epoch & 2) {\small $\sim$} 6.12s per epoch ({\footnotesize $>$} 3x faster)  \\
\bottomrule
\end{tabular}
}
\vspace{-6pt}
\caption{The differences between DGCNN~\cite{wang2019dynamic} proposed by human and PointNet-CSR proposed using \textsc{Dolphin}.}
\label{tab:case_study}
\vspace{-5pt}
\end{table}

We illustrate our approach with an example drawn from 3D point classification task as shown in Fig.~\ref{fig:case_study}. It can be seen that \textsc{Dolphin} can generate codes corresponding to the idea and only in this way, the generated ideas can be effectively verified. 
Tab.~\ref{tab:case_study_sota} presents the comparison between AI-generated approach (\ie, PointNet-CSR obtained by \textsc{Dolphin}) and previous human-designed methods. Idea automatically generated and performed by \textsc{Dolphin} can outperform most human-designed methods and achieve comparable performance to current SoTA~\cite{wang2024gpsformer}. Furthermore, as shown in Tab.~\ref{tab:case_study}, we carefully investigate the existing works on 3D classification task up to the submission date, identifying the work most relevant to PointNet-CSR (AI-generated 3D work), as illustrated in Fig.~\ref{fig:case_study}. The detailed comparison of the idea, implementation, and result can be found in Tab.~\ref{tab:case_study}, showing that PointNet-CSR can achieve better and faster performance through a more concise architecture. Please refer to Fig.~\ref{fig:case_study} and Appendix~\ref{sec:case_studies} for more comparisons between human-designed works and \textsc{Dolphin}-generated works.


\section{Conclusion and Future Outlook}

\textsc{Dolphin} evaluates idea quality through experimental verification, improving it in a closed-loop fashion. Besides, beyond conducting quantitative experiments that demonstrate \textsc{Dolphin}'s capability to generate solutions and results comparable to human-designed approaches, we also conducted in-depth case studies to evaluate the novelty of the ideas and the implementation efficiency of the codes generated by \textsc{Dolphin}. These quantitative and qualitative evaluations we conducted for \textsc{Dolphin} are essential for gaining further insight into the potential and value inherent in \textsc{Dolphin}.

In the future, we envision \textsc{Dolphin} further advancing AI-driven automated scientific research. By harnessing its ability to generate novel ideas in a closed-loop system, we also aspire for \textsc{Dolphin} to foster the development of groundbreaking ideas inspired by cross-disciplinary knowledge, ultimately providing innovative solutions for complex scientific challenges.


\section*{Limitations}

Although \textsc{Dolphin} can generate and implement effective ideas in a loop, it still has inherent limitations. First, in the idea generation stage, LLMs retain historical knowledge obtained from training data, which may cause knowledge leakage when generating ideas. Besides, only the abstract and title are used in the idea generation process, which may result in LLM not being able to understand the technical details in the article and the logic between the articles. However, the feedback-driven idea generation process in \textsc{Dolphin} can be flexibly combined with methods that address these concerns. In addition, the code capabilities of LLMs are not sufficient to understand complex codes such as project-level codes, resulting in \textsc{Dolphin} being unable to verify complex tasks currently.

\section*{Ethics Statement}
This work focuses on the development of a closed-loop framework for auto-research, aiming to accelerate the research cycle. The proposed method relies on publicly available datasets, research papers, and models, ensuring compliance with copyright and intellectual property laws. While the framework is intended to aid human research, we encourage users to critically evaluate the generated results to ensure that they adhere to ethical research practices and mitigate any potential limitations, such as bias or incomplete ideas.

\bibliography{custom}

\begin{thebibliography}{52}
\providecommand{\natexlab}[1]{#1}

\bibitem[{Achiam et~al.(2023)Achiam, Adler, Agarwal, Ahmad, Akkaya, Aleman, Almeida, Altenschmidt, Altman, Anadkat et~al.}]{achiam2023gpt4}
Josh Achiam, Steven Adler, Sandhini Agarwal, Lama Ahmad, Ilge Akkaya, Florencia~Leoni Aleman, Diogo Almeida, Janko Altenschmidt, Sam Altman, Shyamal Anadkat, et~al. 2023.
\newblock Gpt-4 technical report.
\newblock \emph{arXiv preprint arXiv:2303.08774}.

\bibitem[{Anthropic(2024)}]{Anthropic2024claude3}
Anthropic. 2024.
\newblock The claude 3 model family: Opus, sonnet, haiku.
\newblock URL: \url{https://www-cdn.anthropic.com/de8ba9b01c9ab7cbabf5c33b80b7bbc618857627/Model_Card_Claude_3.pdf}.

\bibitem[{Assafelovic(2023)}]{assafelovic2023gptresearcher}
Assafelovic. 2023.
\newblock gpt-researcher.
\newblock URL: \url{https://github.com/assafelovic/gpt-researcher}.

\bibitem[{Chan et~al.(2024)Chan, Chowdhury, Jaffe, Aung, Sherburn, Mays, Starace, Liu, Maksin, Patwardhan et~al.}]{chan2024mle}
Jun~Shern Chan, Neil Chowdhury, Oliver Jaffe, James Aung, Dane Sherburn, Evan Mays, Giulio Starace, Kevin Liu, Leon Maksin, Tejal Patwardhan, et~al. 2024.
\newblock Mle-bench: Evaluating machine learning agents on machine learning engineering.
\newblock \emph{arXiv preprint arXiv:2410.07095}.

\bibitem[{Chen et~al.(2024)Chen, Dohan, and So}]{chen2024evoprompting}
Angelica Chen, David Dohan, and David So. 2024.
\newblock Evoprompting: language models for code-level neural architecture search.
\newblock \emph{Advances in Neural Information Processing Systems}, 36.

\bibitem[{Chen et~al.(2023)Chen, Han, Gong, Bai, Ling, Luo, Chen, Ma, Zhang, Su et~al.}]{chen2023fengwu}
Kang Chen, Tao Han, Junchao Gong, Lei Bai, Fenghua Ling, Jing-Jia Luo, Xi~Chen, Leiming Ma, Tianning Zhang, Rui Su, et~al. 2023.
\newblock Fengwu: Pushing the skillful global medium-range weather forecast beyond 10 days lead.
\newblock \emph{arXiv preprint arXiv:2304.02948}.

\bibitem[{Devlin et~al.(2019)Devlin, Chang, Lee, and Toutanova}]{devlin2019bert}
Jacob Devlin, Ming-Wei Chang, Kenton Lee, and Kristina Toutanova. 2019.
\newblock Bert: Pre-training of deep bidirectional transformers for language understanding.
\newblock In \emph{Proceedings of the 2019 Conference of the North {A}merican Chapter of the Association for Computational Linguistics}, pages 4171--4186.

\bibitem[{Dominik~Schmidt(2024)}]{schmidt2024aide}
Yuxiang~Wu Dominik~Schmidt, Zhengyao~Jiang. 2024.
\newblock Aide.
\newblock URL: \url{https://www.weco.ai/blog/technical-report}.

\bibitem[{Dosovitskiy et~al.(2020)Dosovitskiy, Beyer, Kolesnikov, Weissenborn, Zhai, Unterthiner, Dehghani, Minderer, Heigold, Gelly et~al.}]{dosovitskiy2020image}
Alexey Dosovitskiy, Lucas Beyer, Alexander Kolesnikov, Dirk Weissenborn, Xiaohua Zhai, Thomas Unterthiner, Mostafa Dehghani, Matthias Minderer, Georg Heigold, Sylvain Gelly, et~al. 2020.
\newblock An image is worth 16x16 words: Transformers for image recognition at scale.
\newblock In \emph{International Conference on Learning Representations}.

\bibitem[{Dubey et~al.(2024)Dubey, Jauhri, Pandey, Kadian, Al-Dahle, Letman, Mathur, Schelten, Yang, Fan et~al.}]{dubey2024llama3}
Abhimanyu Dubey, Abhinav Jauhri, Abhinav Pandey, Abhishek Kadian, Ahmad Al-Dahle, Aiesha Letman, Akhil Mathur, Alan Schelten, Amy Yang, Angela Fan, et~al. 2024.
\newblock The llama 3 herd of models.
\newblock \emph{arXiv preprint arXiv:2407.21783}.

\bibitem[{Han et~al.(2024)Han, Chen, Guo, Xu, and Bai}]{han2024cra5}
Tao Han, Zhenghao Chen, Song Guo, Wanghan Xu, and Lei Bai. 2024.
\newblock Cra5: Extreme compression of era5 for portable global climate and weather research via an efficient variational transformer.
\newblock \emph{arXiv preprint arXiv:2405.03376}.

\bibitem[{He et~al.(2016)He, Zhang, Ren, and Sun}]{he2016resnet}
Kaiming He, Xiangyu Zhang, Shaoqing Ren, and Jian Sun. 2016.
\newblock Deep residual learning for image recognition.
\newblock In \emph{Proceedings of the IEEE conference on computer vision and pattern recognition}, pages 770--778.

\bibitem[{Hu et~al.(2024)Hu, Fu, Wang, Wang, Li, Xu, Lu, Jin, Pan, and Lan}]{hu2024nova}
Xiang Hu, Hongyu Fu, Jinge Wang, Yifeng Wang, Zhikun Li, Renjun Xu, Yu~Lu, Yaochu Jin, Lili Pan, and Zhenzhong Lan. 2024.
\newblock Nova: An iterative planning and search approach to enhance novelty and diversity of llm generated ideas.
\newblock \emph{arXiv preprint arXiv:2410.14255}.

\bibitem[{Jumper et~al.(2021)Jumper, Evans, Pritzel, Green, Figurnov, Ronneberger, Tunyasuvunakool, Bates, {\v{Z}}{\'\i}dek, Potapenko et~al.}]{jumper2021highly}
John Jumper, Richard Evans, Alexander Pritzel, Tim Green, Michael Figurnov, Olaf Ronneberger, Kathryn Tunyasuvunakool, Russ Bates, Augustin {\v{Z}}{\'\i}dek, Anna Potapenko, et~al. 2021.
\newblock Highly accurate protein structure prediction with alphafold.
\newblock \emph{nature}, 596(7873):583--589.

\bibitem[{Krizhevsky et~al.(2009)Krizhevsky, Hinton et~al.}]{krizhevsky2009cifar}
Alex Krizhevsky, Geoffrey Hinton, et~al. 2009.
\newblock Learning multiple layers of features from tiny images.

\bibitem[{Li et~al.(2024)Li, Xu, Guo, Zhao, Li, Yuan, Zhang, Jiang, Xin, Dang et~al.}]{li2024chain}
Long Li, Weiwen Xu, Jiayan Guo, Ruochen Zhao, Xinxuan Li, Yuqian Yuan, Boqiang Zhang, Yuming Jiang, Yifei Xin, Ronghao Dang, et~al. 2024.
\newblock Chain of ideas: Revolutionizing research in novel idea development with llm agents.
\newblock \emph{arXiv preprint arXiv:2410.13185}.

\bibitem[{Liu et~al.(2024)Liu, Gao, and Li}]{liu2024large}
Siyi Liu, Chen Gao, and Yong Li. 2024.
\newblock Large language model agent for hyper-parameter optimization.
\newblock \emph{arXiv preprint arXiv:2402.01881}.

\bibitem[{Liu et~al.(2022)Liu, Mao, Wu, Feichtenhofer, Darrell, and Xie}]{liu2022convnet}
Zhuang Liu, Hanzi Mao, Chao-Yuan Wu, Christoph Feichtenhofer, Trevor Darrell, and Saining Xie. 2022.
\newblock A convnet for the 2020s.
\newblock In \emph{Proceedings of the IEEE/CVF conference on computer vision and pattern recognition}, pages 11976--11986.

\bibitem[{Lu et~al.(2024)Lu, Lu, Lange, Foerster, Clune, and Ha}]{AIScientist}
Chris Lu, Cong Lu, Robert~Tjarko Lange, Jakob Foerster, Jeff Clune, and David Ha. 2024.
\newblock {The AI Scientist: Towards Fully Automated Open-Ended Scientific Discovery }.
\newblock \emph{ArXiv}, abs/2408.06292.

\bibitem[{Luo et~al.(2022)Luo, Sun, Xia, Qin, Zhang, Poon, and Liu}]{luo2022biogpt}
Renqian Luo, Liai Sun, Yingce Xia, Tao Qin, Sheng Zhang, Hoifung Poon, and Tie-Yan Liu. 2022.
\newblock Biogpt: generative pre-trained transformer for biomedical text generation and mining.
\newblock \emph{Briefings in bioinformatics}, 23(6):bbac409.

\bibitem[{OpenAI(2024)}]{openai2024gpt4o}
OpenAI. 2024.
\newblock Hello gpt-4o.
\newblock \url{https://openai.com/index/hello-gpt-4o/}.

\bibitem[{Qi et~al.(2023)Qi, Zhang, Li, Tian, Zeng, Chen, and Zhou}]{qi2023large}
Biqing Qi, Kaiyan Zhang, Haoxiang Li, Kai Tian, Sihang Zeng, Zhang-Ren Chen, and Bowen Zhou. 2023.
\newblock Large language models are zero shot hypothesis proposers.
\newblock \emph{arXiv preprint arXiv:2311.05965}.

\bibitem[{Qi et~al.(2024)Qi, Zhang, Tian, Li, Chen, Zeng, Hua, Jinfang, and Zhou}]{qi2024large}
Biqing Qi, Kaiyan Zhang, Kai Tian, Haoxiang Li, Zhang-Ren Chen, Sihang Zeng, Ermo Hua, Hu~Jinfang, and Bowen Zhou. 2024.
\newblock Large language models as biomedical hypothesis generators: A comprehensive evaluation.
\newblock \emph{arXiv preprint arXiv:2407.08940}.

\bibitem[{Qi et~al.(2017{\natexlab{a}})Qi, Su, Mo, and Guibas}]{qi2017pointnet}
Charles~R Qi, Hao Su, Kaichun Mo, and Leonidas~J Guibas. 2017{\natexlab{a}}.
\newblock Pointnet: Deep learning on point sets for 3d classification and segmentation.
\newblock In \emph{Proceedings of the IEEE conference on computer vision and pattern recognition}, pages 652--660.

\bibitem[{Qi et~al.(2017{\natexlab{b}})Qi, Yi, Su, and Guibas}]{qi2017pointnet++}
Charles~Ruizhongtai Qi, Li~Yi, Hao Su, and Leonidas~J Guibas. 2017{\natexlab{b}}.
\newblock Pointnet++: Deep hierarchical feature learning on point sets in a metric space.
\newblock \emph{Advances in neural information processing systems}, 30.

\bibitem[{Qian et~al.(2022)Qian, Li, Peng, Mai, Hammoud, Elhoseiny, and Ghanem}]{qian2022pointnext}
Guocheng Qian, Yuchen Li, Houwen Peng, Jinjie Mai, Hasan Hammoud, Mohamed Elhoseiny, and Bernard Ghanem. 2022.
\newblock Pointnext: Revisiting pointnet++ with improved training and scaling strategies.
\newblock \emph{Advances in neural information processing systems}, 35:23192--23204.

\bibitem[{Qiu et~al.(2023)Qiu, Jiang, Lu, Sclar, Pyatkin, Bhagavatula, Wang, Kim, Choi, Dziri et~al.}]{qiu2023phenomenal}
Linlu Qiu, Liwei Jiang, Ximing Lu, Melanie Sclar, Valentina Pyatkin, Chandra Bhagavatula, Bailin Wang, Yoon Kim, Yejin Choi, Nouha Dziri, et~al. 2023.
\newblock Phenomenal yet puzzling: Testing inductive reasoning capabilities of language models with hypothesis refinement.
\newblock \emph{arXiv preprint arXiv:2310.08559}.

\bibitem[{Randolph(2019)}]{randolph2019guide}
Justus Randolph. 2019.
\newblock A guide to writing the dissertation literature review.
\newblock \emph{Practical assessment, research, and evaluation}, 14(1):13.

\bibitem[{Reimers and Gurevych(2020)}]{reimers-2020-multilingual-sentence-bert}
Nils Reimers and Iryna Gurevych. 2020.
\newblock {Making Monolingual Sentence Embeddings Multilingual using Knowledge Distillation}.
\newblock In \emph{EMNLP}.

\bibitem[{Schmidgall et~al.(2025)Schmidgall, Su, Wang, Sun, Wu, Yu, Liu, Liu, and Barsoum}]{schmidgall2025agent}
Samuel Schmidgall, Yusheng Su, Ze~Wang, Ximeng Sun, Jialian Wu, Xiaodong Yu, Jiang Liu, Zicheng Liu, and Emad Barsoum. 2025.
\newblock Agent laboratory: Using llm agents as research assistants.
\newblock \emph{arXiv preprint arXiv:2501.04227}.

\bibitem[{Si et~al.(2024)Si, Yang, and Hashimoto}]{si2024stanfordcan}
Chenglei Si, Diyi Yang, and Tatsunori Hashimoto. 2024.
\newblock Can llms generate novel research ideas? a large-scale human study with 100+ nlp researchers.
\newblock \emph{arXiv preprint arXiv:2409.04109}.

\bibitem[{Socher et~al.(2013)Socher, Perelygin, Wu, Chuang, Manning, Ng, and Potts}]{socher2013recursive}
Richard Socher, Alex Perelygin, Jean Wu, Jason Chuang, Christopher~D Manning, Andrew~Y Ng, and Christopher Potts. 2013.
\newblock Recursive deep models for semantic compositionality over a sentiment treebank.
\newblock In \emph{Proceedings of the 2013 conference on empirical methods in natural language processing}, pages 1631--1642.

\bibitem[{Su et~al.(2024)Su, Chen, Tang, Zheng, Li, Yin, Ouyang, and Dong}]{su2024two}
Haoyang Su, Renqi Chen, Shixiang Tang, Xinzhe Zheng, Jingzhe Li, Zhenfei Yin, Wanli Ouyang, and Nanqing Dong. 2024.
\newblock Two heads are better than one: A multi-agent system has the potential to improve scientific idea generation.
\newblock \emph{arXiv preprint arXiv:2410.09403}.

\bibitem[{Team(2023{\natexlab{a}})}]{copilot}
Copilot Team. 2023{\natexlab{a}}.
\newblock copilot.
\newblock URL: \url{https://github.com/features/copilot}.

\bibitem[{Team(2023{\natexlab{b}})}]{ollama}
Ollama Team. 2023{\natexlab{b}}.
\newblock ollama.
\newblock URL: \url{https://ollama.com/}.

\bibitem[{Wang et~al.(2024{\natexlab{a}})Wang, Wu, Lam, Ning, Yu, Wang, Li, and Srikanthan}]{wang2024gpsformer}
Changshuo Wang, Meiqing Wu, Siew-Kei Lam, Xin Ning, Shangshu Yu, Ruiping Wang, Weijun Li, and Thambipillai Srikanthan. 2024{\natexlab{a}}.
\newblock Gpsformer: A global perception and local structure fitting-based transformer for point cloud understanding.
\newblock In \emph{European Conference on Computer Vision}. Springer.

\bibitem[{Wang(2023)}]{wang2023octformer}
Peng-Shuai Wang. 2023.
\newblock Octformer: Octree-based transformers for 3d point clouds.
\newblock \emph{ACM Transactions on Graphics (TOG)}, 42(4):1--11.

\bibitem[{Wang et~al.(2023)Wang, Downey, Ji, and Hope}]{wang2023scimon}
Qingyun Wang, Doug Downey, Heng Ji, and Tom Hope. 2023.
\newblock Scimon: Scientific inspiration machines optimized for novelty.
\newblock \emph{arXiv preprint arXiv:2305.14259}.

\bibitem[{Wang et~al.(2024{\natexlab{b}})Wang, Downey, Ji, and Hope}]{Wang2023SciMONSI}
Qingyun Wang, Doug Downey, Heng Ji, and Tom Hope. 2024{\natexlab{b}}.
\newblock {SciMON: Scientific Inspiration Machines Optimized for Novelty}.
\newblock In \emph{ACL}.

\bibitem[{Wang et~al.(2024{\natexlab{c}})Wang, Guo, Yao, Zhang, Zhang, Wu, Zhang, Dai, Zhang, Wen et~al.}]{wang2024autosurvey}
Yidong Wang, Qi~Guo, Wenjin Yao, Hongbo Zhang, Xin Zhang, Zhen Wu, Meishan Zhang, Xinyu Dai, Min Zhang, Qingsong Wen, et~al. 2024{\natexlab{c}}.
\newblock Autosurvey: Large language models can automatically write surveys.
\newblock \emph{arXiv preprint arXiv:2406.10252}.

\bibitem[{Wang et~al.(2019)Wang, Sun, Liu, Sarma, Bronstein, and Solomon}]{wang2019dynamic}
Yue Wang, Yongbin Sun, Ziwei Liu, Sanjay~E Sarma, Michael~M Bronstein, and Justin~M Solomon. 2019.
\newblock Dynamic graph cnn for learning on point clouds.
\newblock \emph{ACM Transactions on Graphics (tog)}, 38(5):1--12.

\bibitem[{Wu et~al.(2015)Wu, Song, Khosla, Yu, Zhang, Tang, and Xiao}]{wu2015modelnet}
Zhirong Wu, Shuran Song, Aditya Khosla, Fisher Yu, Linguang Zhang, Xiaoou Tang, and Jianxiong Xiao. 2015.
\newblock 3d shapenets: A deep representation for volumetric shapes.
\newblock In \emph{Proceedings of the IEEE conference on computer vision and pattern recognition}, pages 1912--1920.

\bibitem[{Yan et~al.(2025)Yan, Feng, Yuan, Xia, Wang, Zhang, and Bai}]{yan2025surveyforge}
Xiangchao Yan, Shiyang Feng, Jiakang Yuan, Renqiu Xia, Bin Wang, Bo~Zhang, and Lei Bai. 2025.
\newblock Surveyforge: On the outline heuristics, memory-driven generation, and multi-dimensional evaluation for automated survey writing.
\newblock \emph{arXiv preprint arXiv:2503.04629}.

\bibitem[{Yang et~al.(2024{\natexlab{a}})Yang, Yang, Hui, Zheng, Yu, Zhou, Li, Li, Liu, Huang et~al.}]{yang2024qwen2}
An~Yang, Baosong Yang, Binyuan Hui, Bo~Zheng, Bowen Yu, Chang Zhou, Chengpeng Li, Chengyuan Li, Dayiheng Liu, Fei Huang, et~al. 2024{\natexlab{a}}.
\newblock Qwen2 technical report.
\newblock \emph{arXiv preprint arXiv:2407.10671}.

\bibitem[{Yang et~al.(2024{\natexlab{b}})Yang, Du, Li, Zheng, Poria, and Cambria}]{Yang2023LargeLM}
Zonglin Yang, Xinya Du, Junxian Li, Jie Zheng, Soujanya Poria, and E.~Cambria. 2024{\natexlab{b}}.
\newblock {Large Language Models for Automated Open-domain Scientific Hypotheses Discovery}.
\newblock \emph{ACL Findings}.

\bibitem[{Yang et~al.(2023)Yang, Du, Li, Zheng, Poria, and Cambria}]{yang2023large}
Zonglin Yang, Xinya Du, Junxian Li, Jie Zheng, Soujanya Poria, and Erik Cambria. 2023.
\newblock Large language models for automated open-domain scientific hypotheses discovery.
\newblock \emph{arXiv preprint arXiv:2309.02726}.

\bibitem[{Yao et~al.(2024)Yao, Ke, Wang, Li, and Hu}]{yao2024lawyer}
Shunyu Yao, Qingqing Ke, Qiwei Wang, Kangtong Li, and Jie Hu. 2024.
\newblock \href {https://doi.org/10.1145/3689299.3689319} {Lawyer gpt: A legal large language model with enhanced domain knowledge and reasoning capabilities}.
\newblock In \emph{Proceedings of the 2024 3rd International Symposium on Robotics, Artificial Intelligence and Information Engineering}, RAIIE '24, page 108–112, New York, NY, USA. Association for Computing Machinery.

\bibitem[{Zagoruyko(2016)}]{zagoruyko2016wrn}
Sergey Zagoruyko. 2016.
\newblock Wide residual networks.
\newblock \emph{arXiv preprint arXiv:1605.07146}.

\bibitem[{Zhang et~al.(2023{\natexlab{a}})Zhang, Zhang, Ren, Li, and Yang}]{zhang2023mlcopilot}
Lei Zhang, Yuge Zhang, Kan Ren, Dongsheng Li, and Yuqing Yang. 2023{\natexlab{a}}.
\newblock Mlcopilot: Unleashing the power of large language models in solving machine learning tasks.
\newblock \emph{arXiv preprint arXiv:2304.14979}.

\bibitem[{Zhang et~al.(2023{\natexlab{b}})Zhang, Gong, Wu, Liu, and Zhou}]{zhang2023automl}
Shujian Zhang, Chengyue Gong, Lemeng Wu, Xingchao Liu, and Mingyuan Zhou. 2023{\natexlab{b}}.
\newblock Automl-gpt: Automatic machine learning with gpt.
\newblock \emph{arXiv preprint arXiv:2305.02499}.

\bibitem[{Zhou et~al.(2024)Zhou, Liu, Srivastava, Mei, and Tan}]{zhou2024hypothesis}
Yangqiaoyu Zhou, Haokun Liu, Tejes Srivastava, Hongyuan Mei, and Chenhao Tan. 2024.
\newblock Hypothesis generation with large language models.
\newblock \emph{arXiv preprint arXiv:2404.04326}.

\bibitem[{Zhu et~al.(2024)Zhu, Guo, Shao, Yang, Wang, Xu, Wu, Li, Gao, Ma et~al.}]{zhu2024deepseek}
Qihao Zhu, Daya Guo, Zhihong Shao, Dejian Yang, Peiyi Wang, Runxin Xu, Y~Wu, Yukun Li, Huazuo Gao, Shirong Ma, et~al. 2024.
\newblock Deepseek-coder-v2: Breaking the barrier of closed-source models in code intelligence.
\newblock \emph{arXiv preprint arXiv:2406.11931}.

\end{thebibliography}

\appendix

\clearpage
\appendix

\section*{Outlines for Appendix}
\label{sec:appendix}

In Appendix, we provide additional details and qualitative results on the following aspects:

\begin{itemize}
    \item Appendix~\ref{sec:dolphin_prompt}: Further details of \textsc{Dolphin} including prompts, qualitative results for each step, and so on:
     \begin{itemize}
        \item Appendix~\ref{ideas}: Ideas generation process.
        \item Appendix~\ref{experimental}: Experimental verification process.
    \end{itemize}
    \item Appendix~\ref{sec:exp}: Additional details and results of performing experiments using \textsc{Dolphin}:
     \begin{itemize}
        \item Appendix~\ref{task} The selected datasets and implementation on different tasks:
        \begin{itemize}
            \item Appendix.~\ref{img_cls}: Image classification task.
            \item Appendix~\ref{3d_cls}: 3D classification task.
            \item Appendix~\ref{sentiment_cls}: Sentiment classification task.
            \item Appendix~\ref{mlebench}: MLE-bench.
        \end{itemize}
        \item Appendix~\ref{qualitative}: Code implementation differences between human-designed and Dolphin-designed.
     \end{itemize}
    \item Appendix~\ref{sec:analysis}: More analyses and further works:
    \begin{itemize}
        \item Appendix~\ref{extendability}: The extensibility of the research task related to idea generation.
        \item Appendix~\ref{future_works}: Analyses on future works.
    \end{itemize}

\end{itemize}

\section{Further details of \textbf{Dolphin}}
\label{sec:dolphin_prompt}

In this section, we provide more details about \textsc{Dolphin} including prompts, qualitative results of some processes, and algorithms of some processes. In the following section, we will give more details of the ideas generation process, experimental verification process, and results feedback process, respectively.

\subsection{Ideas Generation Process}
\label{ideas}

We provide prompts used in paper retrieval, paper ranking, and idea generation process in Fig.~\ref{fig:prompt_paper}. We partially refer to the prompt design of previous works~\cite{AIScientist,si2024stanfordcan}. As depicted in our manuscript, the quality of retrieved papers is important to idea generation. Here, we give an example to further illustrate the impact. Given the topic \textit{`3D classification'}, naive retrieval will result in lots of papers related to 3D object detection. As a result, we have identified several ideas that are more closely related to detection (\eg, region proposal PointNet), which are significantly influenced by the detection task.

Further, we provide the algorithm of independence check in Algorithm~\ref{independence_alg}. To show the effectiveness of the independence check process, we show an example in Fig.~\ref{fig:example_ind}. It can be seen that although the name and title of the idea are totally different from each other, the technologies used in the two ideas are almost the same. Our idea independence check process can effectively filter the repeated ideas, further improving the auto-research efficiency.

\setcounter{figure}{4}
\begin{figure*}[h]
\vspace{-10pt}
  \centering
   \includegraphics[width=0.88\linewidth]{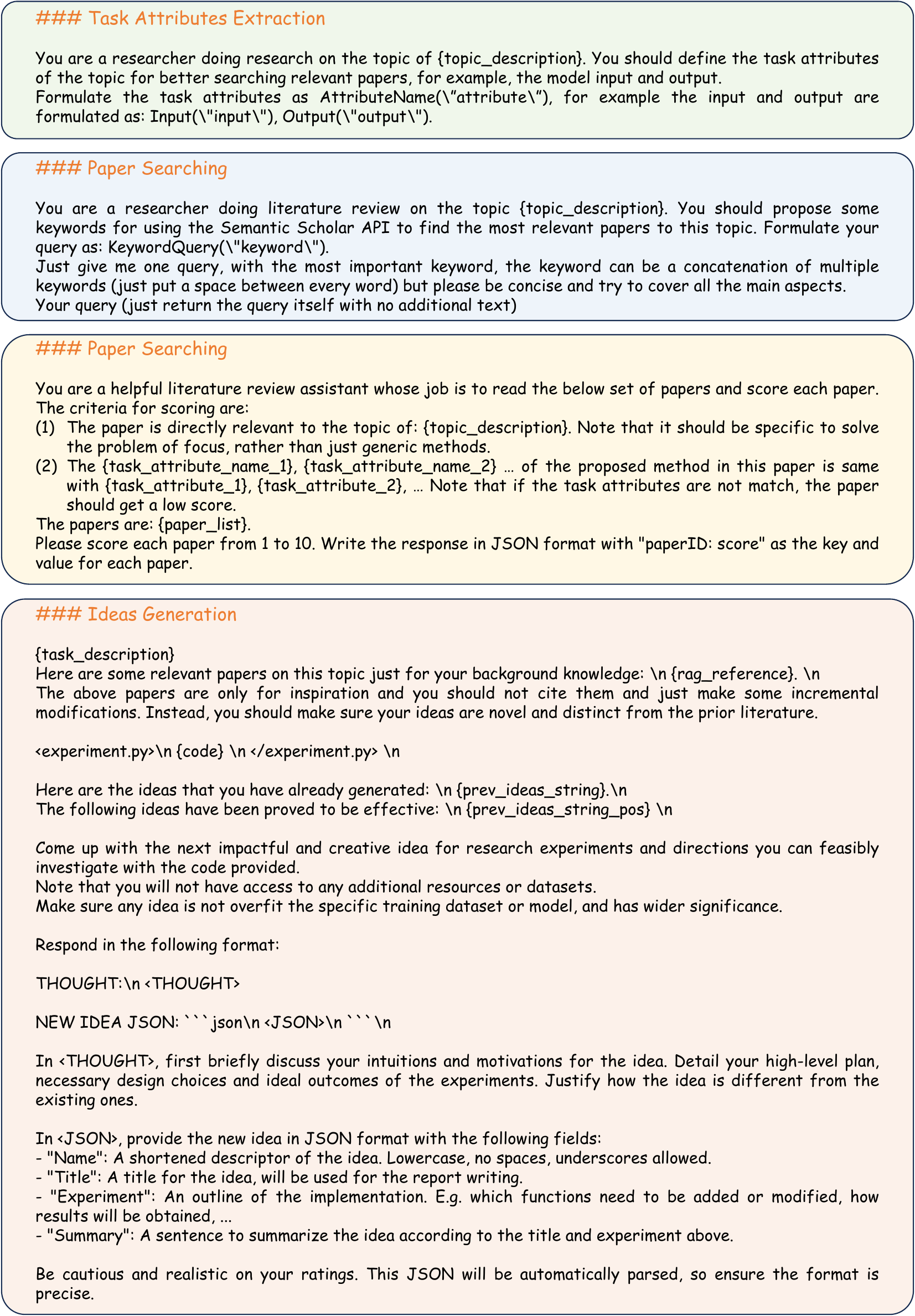}
   \vspace{-4pt}
   \caption{Prompts of paper retrieval, paper ranking, and ideas generation.}
   \label{fig:prompt_paper}
\end{figure*}

\begin{figure*}[h!]
\vspace{-10pt}
  \centering
   \includegraphics[width=0.90\linewidth]{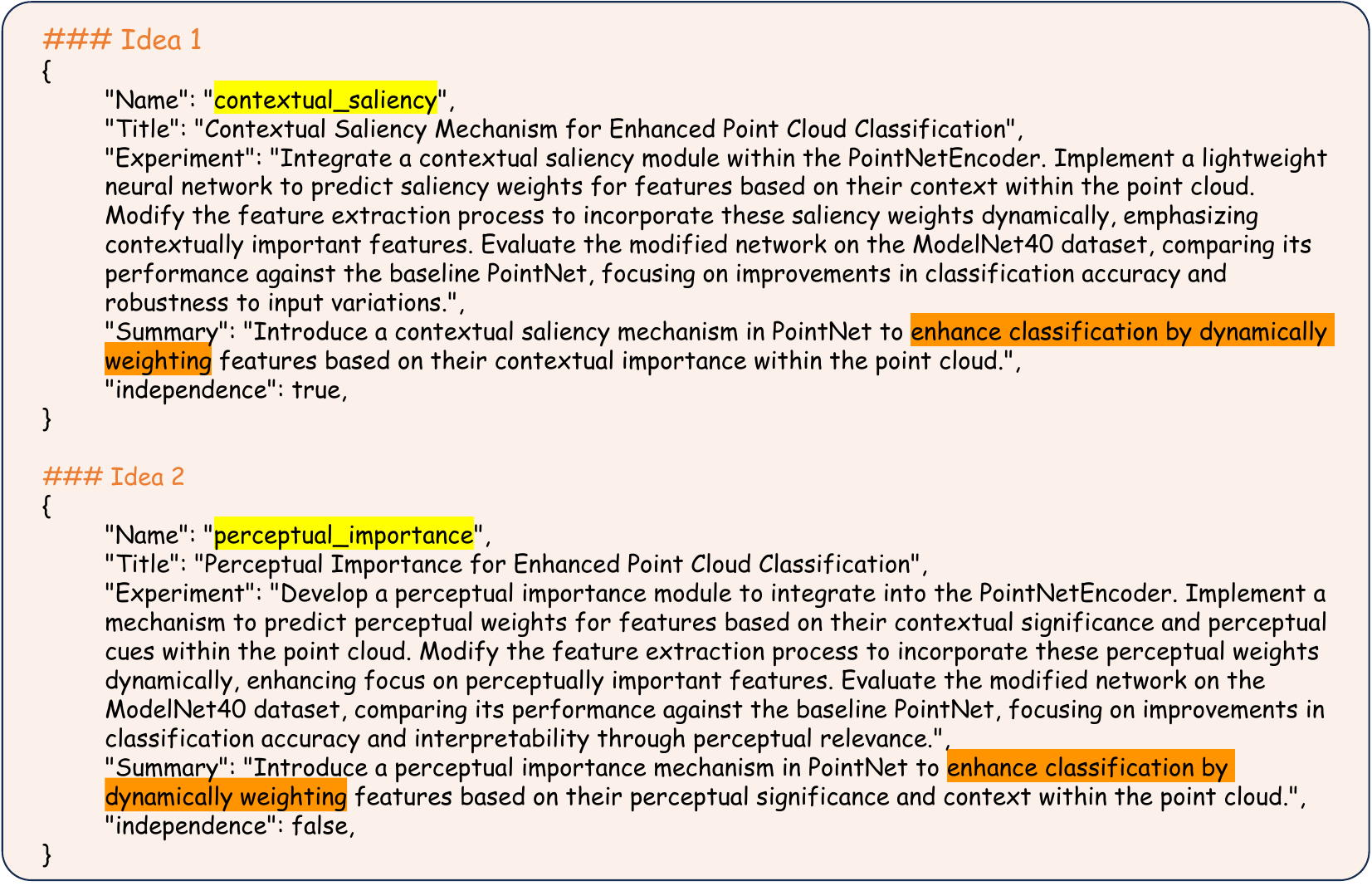}
   \vspace{-4pt}
   \caption{An example of independence check.}
   \label{fig:example_ind}
\end{figure*}

\begin{algorithm}[t]
\caption{Independence Check Process}
\label{independence_alg}
\KwIn{List of ideas $\mathbf{I}$, previous paper memory bank $\mathbf{B}$, sentence embedding model $\mathcal{S}$, threshold $\tau$.}
\KwOut{Idea independence of each paper $\mathbf{R}$.}
\For{idea summary $s$ in $\mathbf{I}$}{
    \If{$\text{len}(\mathbf{B}$) == 0}{
        $\mathbf{R}$.append(True)
    }
    \Else{
        Extract summary embedding: $f_s=\mathcal{S}(s)$. \\
        Compare $f_s$ with summary embeddings in $\mathbf{B}$ by $sim=f_s \cdot \mathbf{B}^T \in \mathbf{R}^{1 \times len(\mathbf{B})}$. \\
        \If{$\max(sim)<\tau$}{
            $\mathbf{R}$.append(True)
        }
        \Else{
            $\mathbf{R}$.append(False)
        }
    }
}

\Return{Independence list $\mathbf{R}$, len($\mathbf{R}$) == len($\mathbf{I}$)}
\end{algorithm}

\subsection{Experimental Verification Process}
\label{experimental}

\begin{figure*}[h]
\vspace{-10pt}
  \centering
   \includegraphics[width=0.90\linewidth]{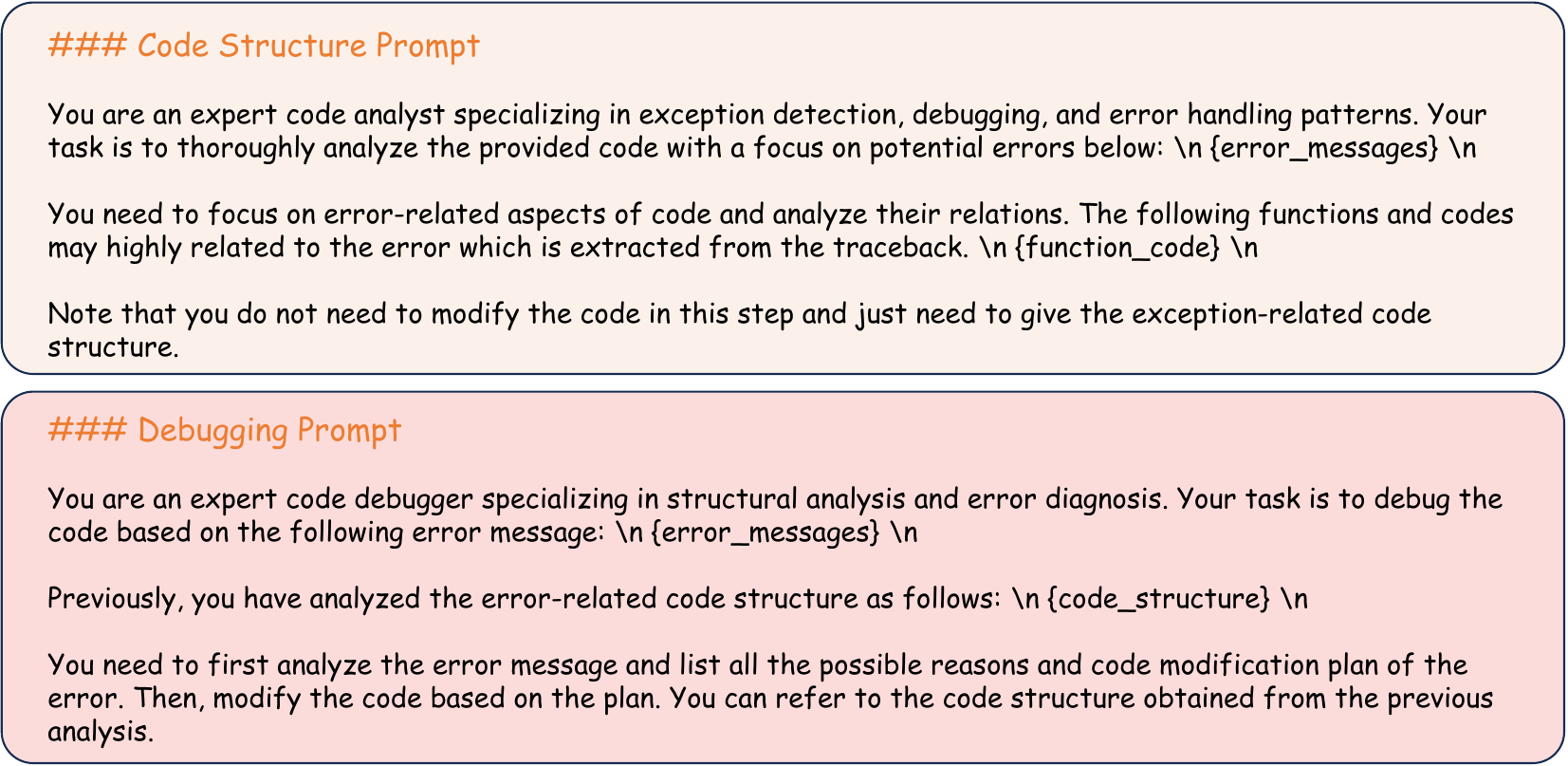}
   \vspace{-4pt}
   \caption{Prompts of local code structure and debugging.}
   \label{fig:code_prompt}
\end{figure*}

\begin{figure*}[h]
\vspace{-10pt}
  \centering
   \includegraphics[width=0.95\linewidth]{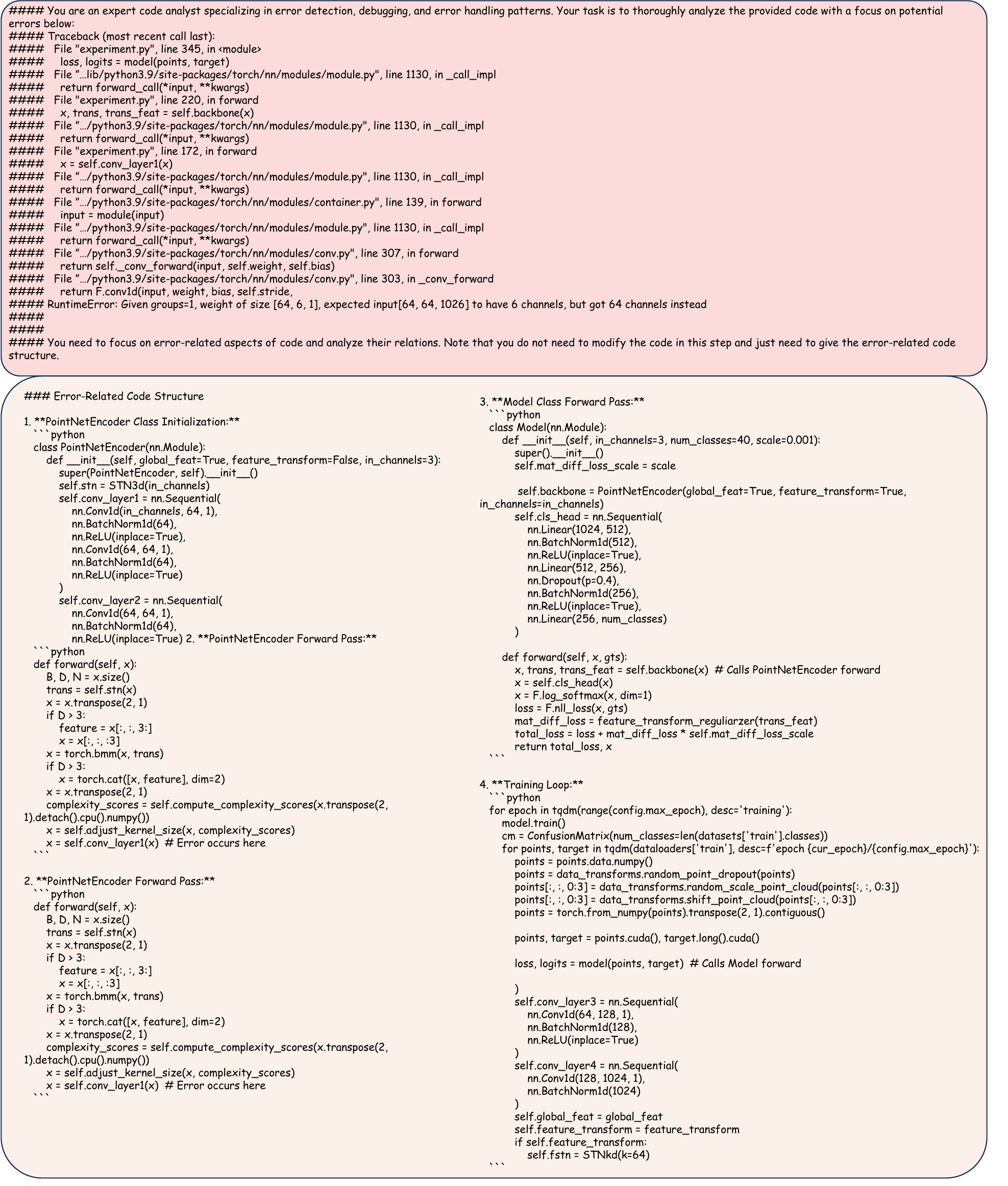}
   \vspace{-4pt}
   \caption{Code structure without extracted traceback information.}
   \label{fig:wo_trace}
\end{figure*}

Fig.~\ref{fig:code_prompt} shows the local code structure generation prompt, which needs first to extract the exception traceback information. Further, to show how this information can guide the LLM in generating the local code structure, we provide an example for better illustration. As shown in Fig.~\ref{fig:wo_trace}, the LLM tend to copy the original reference code which is useless in the following debugging process.

\section{Additional Details and Results of Experiments}
\label{sec:exp}

In this section, we provide a comprehensive overview of the implementation details and dataset information used in our main text.

\subsection{Details of Selected Tasks}
\label{task}

\subsubsection{Image Classification Task}
\label{img_cls}

\noindent \textbf{Dataset: CIFAR-100.} The CIFAR-100 dataset~\cite{krizhevsky2009cifar} includes colored
natural images with a resolution of 32×32 pixels, categorized into 100 distinct classes. The training and test sets contain 50,000 and
10,000 images, respectively. We adopt a standard data augmentation scheme (\ie, RandomCrop, RandomHorizontalFlip, RandomRotation).

\noindent\textbf{Implementation Details.} We use WRN-28-10~\cite{zagoruyko2016wrn} as our baseline. We partially refer to the codebase\footnote{\url{https://github.com/bmsookim/wide-resnet.pytorch}}. Our training process employs the SGD optimizer with the CosineAnuealing scheduler. The initial learning rate is set to 0.1, and we train the model for 200 epochs with a batch size of 128. We apply a weight decay of 5e-4 and a Nesterov momentum with a coefficient of 0.9.

\subsubsection{3D Classification Task}
\label{3d_cls}

\noindent \textbf{Dataset: ModelNet40.} ModelNet40~\cite{wu2015modelnet} is a synthetic object dataset that contains 12,311 3D CAD models covering 40 categories. The standard training/validation set of ModelNet40 carries 9843/2468 point clouds.

\noindent \textbf{Implementation Details.} We use PointNet~\cite{qi2017pointnet} as our baseline. Following PointNet~\cite{qi2017pointnet}, we uniformly sample 1024 points on each object. We use the random scale, random dropout, and point shift during training and train the model for 200 epochs. The initial learning rate is set to 1e-3. We use Adam optimizer (weight decay=1e-4) and step learning rate decay (step size=20, gamma=0.7). Our implementation partially refers to codebase\footnote{\url{https://github.com/yanx27/Pointnet_Pointnet2_pytorch/tree/master}}.

\subsubsection{Sentiment Classification Task}
\label{sentiment_cls}

\noindent \textbf{Dataset: SST-2.} Stanford Sentiment Treebank (SST)~\cite{socher2013recursive} contains 11,855 one-sentence movie
reviews extracted from Rotten Tomatoes. SST contains
215,154 unique manually labeled texts of varying lengths.

\noindent \textbf{Implementation Details.} Our code of sentiment classification tasks refers to the codebase\footnote{\url{https://github.com/YJiangcm/SST-2-sentiment-analysis}}. We fully fine-tune the BERT-base model for the classification task for 5 epochs with the learning rate 2e-5. The batch size is set to 32. We use the early stop mechanism during training (\ie., stop training if the accuracy of current epoch is lower than that of the previous epoch).

\subsubsection{MLE-bench}
\label{mlebench}

MLE-bench~\cite{chan2024mle} is designed to evaluate the capability of AI agents' ML engineering. It is composed of 75 ML competitions from Kaggle. In our manuscript, we conduct experiment on 3 tasks in MLE-bench including detecting insults in social commentary challenge, tabular playground series dec 2021 challenge, and jigsaw toxic comment classification challenge.

\subsection{Case Studies}
\label{sec:case_studies}

We provide several cases that are automatically generated and evaluated by \textsc{Dolphin} as shown in Fig.~\ref{fig:case_1}, Fig.~\ref{fig:case_2}, and Fig.~\ref{fig:case_3}. We show the ideas and modified codes in figures and the performance in the corresponding caption, respectively.

\label{qualitative}

\begin{figure*}[t]
\vspace{-8pt}
  \centering
   \includegraphics[width=1.0\linewidth]{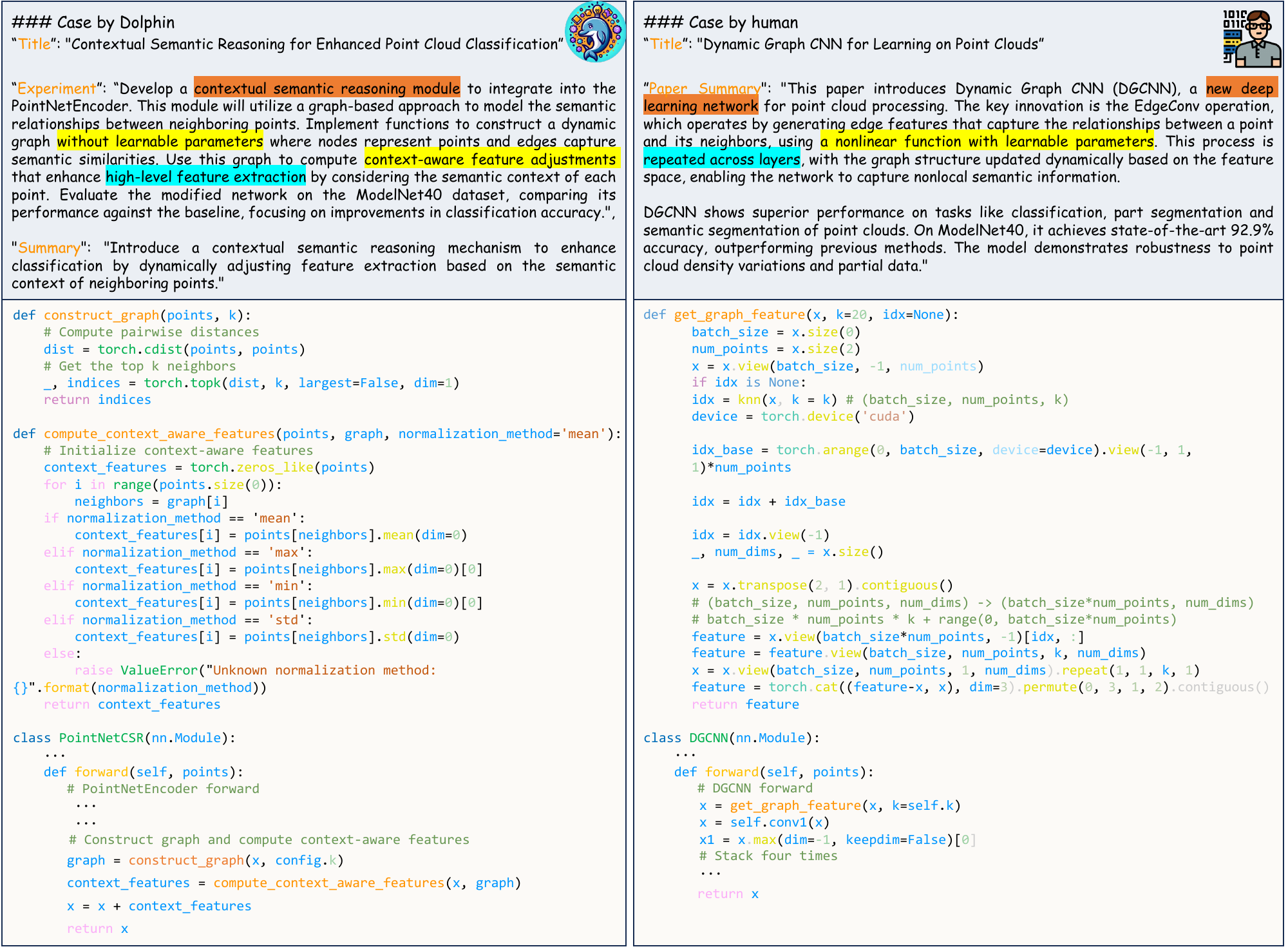}
   \vspace{-6pt}
   \caption{Case studies for the ideas and codes generated by \textsc{Dolphin} (Left) and human researcher (Right).}
   \label{fig:case_study}
\vspace{-2pt}
\end{figure*}

\begin{figure*}[h]
\vspace{-10pt}
  \centering
   \includegraphics[width=1.0\linewidth]{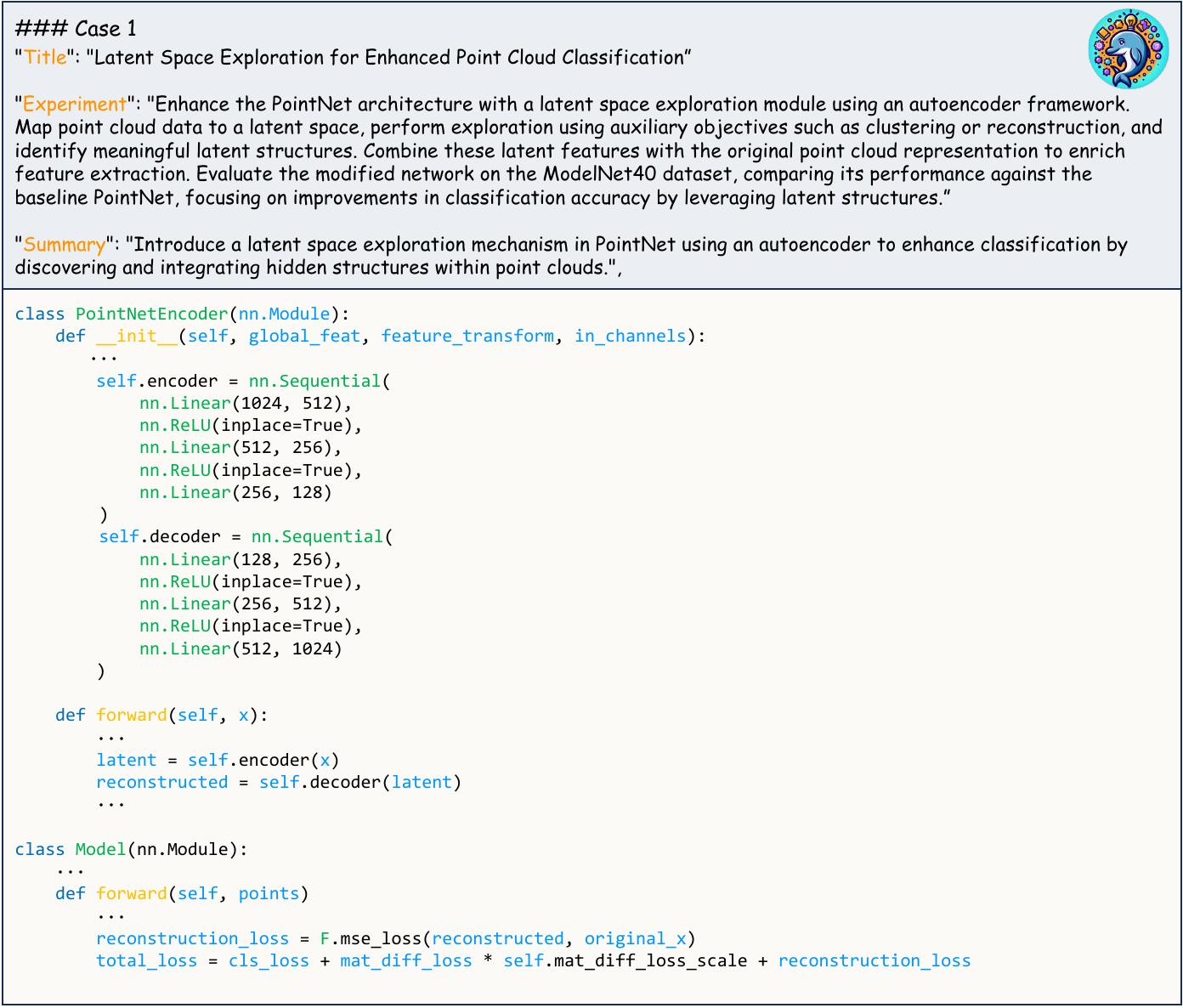}
   \vspace{-4pt}
   \caption{Idea and codes generated by \textsc{Dolphin} which achieves 92.34\% OA and 89.54\% mAcc. on ModelNet40 (+1.34\% OA and +1.94\% mAcc. compared to our baseline).}
   \label{fig:case_1}
\end{figure*}

\begin{figure*}[h]
\vspace{-10pt}
  \centering
   \includegraphics[width=1.0\linewidth]{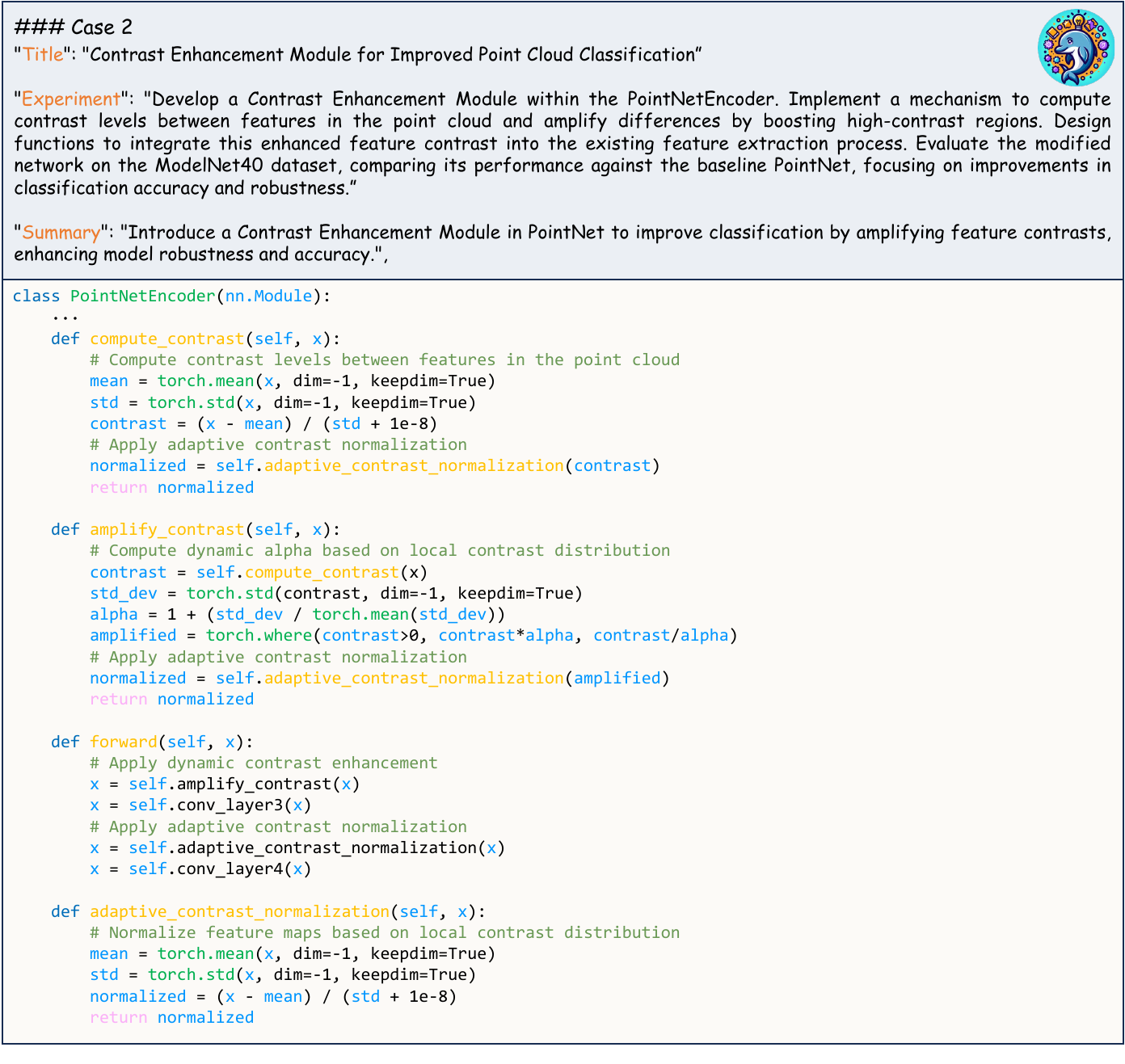}
   \vspace{-4pt}
   \caption{Idea and codes generated by \textsc{Dolphin} which achieves 92.30\% OA and 88.96\% mAcc. on ModelNet40 (+1.30\% OA and +1.36\% mAcc. compared to our baseline).}
   \label{fig:case_2}
\end{figure*}

\begin{figure*}[h]
\vspace{-10pt}
  \centering
   \includegraphics[width=1.0\linewidth]{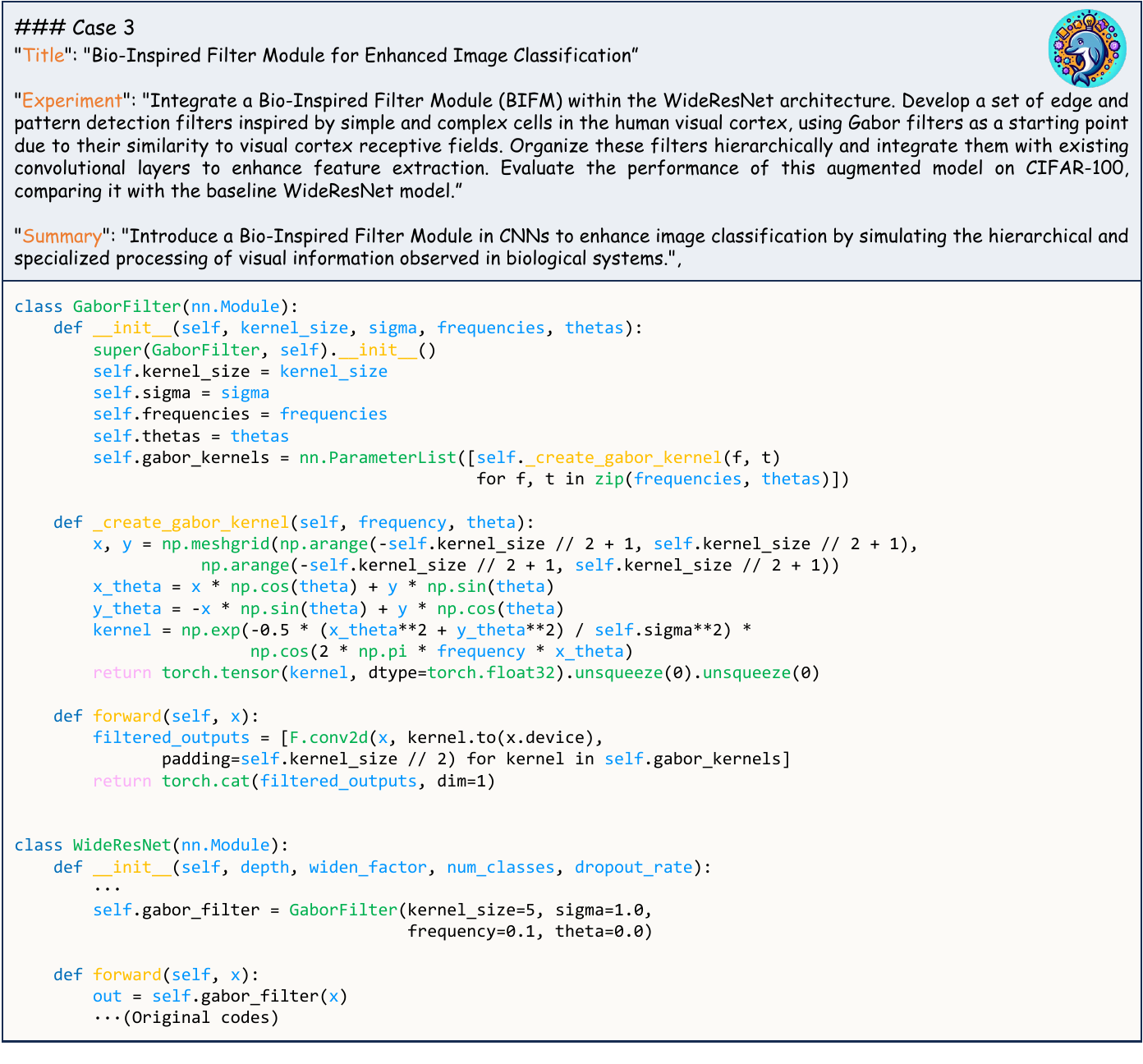}
   \vspace{-4pt}
   \caption{Idea and codes generated by \textsc{Dolphin} which achieves 82.05\% Acc. on CIFAR-100 (+0.85\% Acc. compared to our baseline).}
   \label{fig:case_3}
\end{figure*}

\section{Further Analysis and Future Work}
\label{sec:analysis}

\subsection{The Extensibility for the Research Task}
\label{extendability}

Recently, lots of works have explored automatic idea generation~\cite{li2024chain,si2024stanfordcan}. One limitation is that the novelty of an idea can only be judged through human scoring or large model scoring. However, in real scientific research, we need ideas that can lead to performance breakthroughs that can not be judged without experiments. \textsc{Dolphin} which includes ideas generation, experimental verification, and results feedback process can serve as an evaluation protocol. In the future, it can be combined with auto-idea generation works to assess the effectiveness of the idea generation.

\subsection{Analysis on Future Works}
\label{future_works}
\textsc{Dopline} achieves the first closed-loop automatic research framework, we still hope that \textsc{Dopline} will possess stronger auto-research capabilities. For example, our ultimate goal is to utilize the capabilities of large models to integrate multi-disciplinary knowledge which is hard to be achieved by human researchers. To achieve this goal, we still need to make efforts in the following aspects: \textbf{1)} develop more powerful code models that can understand and modify project-level code, and \textbf{2)} retrieve multi-disciplinary papers that may be related to the given topic.

\end{document}